\documentclass[runningheads]{llncs}
\usepackage[square,sort&compress,comma,numbers]{natbib}
\usepackage[T1]{fontenc}
\usepackage{xcolor}
\usepackage{subfigure}
\usepackage{bbding}
\usepackage{epsfig}
\usepackage{graphicx}
\usepackage{booktabs}       % professional-quality tables
\usepackage{amsmath}
\usepackage{mathtools}
\usepackage{gensymb}
\usepackage{booktabs}       % professional-quality tables
\usepackage{amsfonts}       % blackboard math symbols     
\usepackage{nicefrac}       % compact symbols for 1/2, etc.
\usepackage{microtype}      % microtypography
\usepackage{algorithm}
\usepackage{algorithmic}
\usepackage{mathtools}
\usepackage{hyperref}

\newcommand{\Figure}[1]{Fig.~\ref{#1}}
\newcommand{\Table}[1]{Table~\ref{#1}}

\makeatletter
\newcommand{\mypm}{\mathbin{\mathpalette\@mypm\relax}}
\newcommand{\@mypm}[2]{\ooalign{%
  \raisebox{.1\height}{$#1+$}\cr
  \smash{\raisebox{-.6\height}{$#1-$}}\cr}}
\makeatother

\begin{document}
%
%Realistic Autonomous Vehicle Collision Scenario Generation for Multi Agent Case
%Varied Realistic Autonomous Vehicle Collision Scenario Generation
\title{Semantic and Articulated Pedestrian Sensing Onboard a Moving Vehicle}
% maybe mention varied pedestrian behaviours???
%\thanks{Supported by organization x.}}

% AUTHORS AND INSTITUTIONS COMMENTED OUT FOR ANONYMITY
\author{Maria Priisalu\inst{1}}%\orcidID{0000-1111-2222-3333} \and
%Second Author\inst{2,3}\orcidID{1111-2222-3333-4444} \and
%Third Author\inst{3}\orcidID{2222--3333-4444-5555}}
%%
\authorrunning{M. Priisalu}
%% First names are abbreviated in the running head.
%% If there are more than two authors, 'et al.' is used.
%%
\institute{Lund University, Sweden\\\email{maria.priisalu@math.lu.se}}
%Springer Heidelberg, Tiergartenstr. 17, 69121 Heidelberg, Germany
\maketitle              % typeset the header of the contribution
\begin{abstract}
It is difficult to perform 3D reconstruction from on-vehicle gathered video due to the large forward motion of the vehicle. Even object detection and human sensing models perform significantly worse on onboard videos when compared to standard benchmarks because objects often appear far away from the camera compared to the standard object detection benchmarks, image quality is often decreased by motion blur and occlusions occur often. This has led to the popularisation of traffic data-specific benchmarks. Recently Light
Detection And Ranging (LiDAR) sensors have become popular to directly estimate depths without the need to perform 3D reconstructions. However, LiDAR-based methods still lack in articulated human detection at a distance when compared to image-based methods. We hypothesize that benchmarks targeted at articulated human sensing from LiDAR data could bring about increased research in human sensing and prediction in traffic and could lead to improved traffic safety for pedestrians.
\end{abstract}
%
%
% To test & develop AV need to be able to model the world from data (as completely as possible).
% Human collisions should be avoided because lethal
% More research on sensing humans in images, but proposed methodology could be improved with other sensors (LIDAR used in my paper).
% Goal: reconstruct 3D scene with semantics & articulated pedestrians.
% This allows for 
% - new data generation
% - external traffic modelling
% - AV modelling & testing.
% Compare to related work
% Present pipeline
\section{Introduction}
\begin{figure}
\centering
\vspace{-0.5cm}
\includegraphics[width=\textwidth]{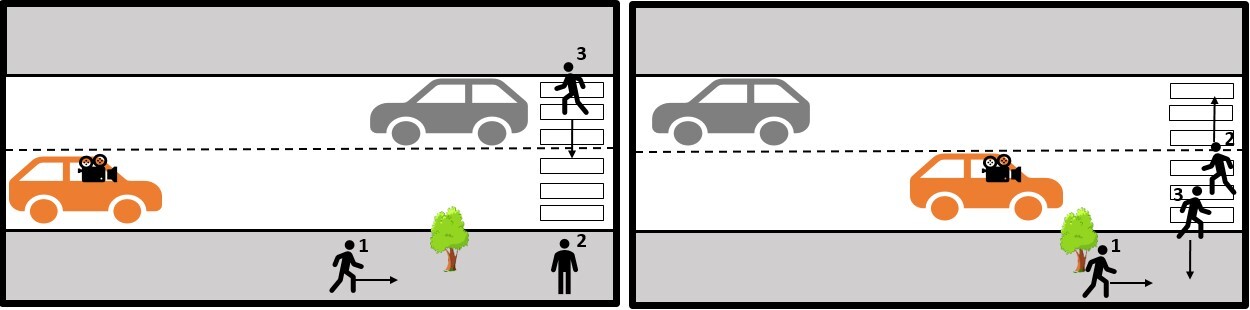}
\vspace{-0.8cm}
\caption{By semantically modeling articulated pedestrians an AV in orange in the left figure can foresee that pedestrian 1 will continue moving in the same direction eventually being occluded by the tree (see right figure), that pedestrian 2 may choose to cross when standing next to a crosswalk (see right figure), and that the third pedestrian will continue to cross once visible. Modeling articulated pedestrians will also ease the AV to differentiate between the second and third pedestrians as their paths cross (see figure on right), as a sudden change in direction is unlikely on a crossroad and given the pedestrians' articulated pose.} \label{problem}
\vspace{-0.4cm}
\end{figure}

Autonomous vehicle (AV) research is gaining momentum ~\cite{paden2016survey,zhu2021survey,ye2021survey,WeiLGGL21} in modeling vehicle-to-vehicle interactions, but  pedestrian-vehicle motion planning models~\cite{bae2023set,shen2018transferable,li2020learning,rhinehart2019precog,yao2021bitrap,huang2020long,deo2020trajectory,zhao2021tnt,mangalam2020not,li2022evolvehypergraph,chen2022hierarchical,djuric2020uncertainty,liang2020pnpnet,luo2022gamma,ma2019trafficpredict,zhu2020robust,sriram2020smart,zhao2019multi,luo2021safety,fang2020tpnet,park2020diverse,vansafecritic,yang2020traffic,cheng2021amenet,giuliari2021transformer,AndersonVJ20a,SalzmannICP20,HamandiDF19,Yao2019FollowingSG,ChenLSL20,IvanovicLSP21,GirgisGCWDKHP22,tang2023collaborative,huang2020diversitygan,jiang2022perceive,tang2019multiple,manish2023survey,gilles2022thomas,li2020end,zeng2020deep,casas2020spagnn,messaoud2021trajectory} could be improved by articulated human motion modelling. Pedestrians in difference to vehicles provide strong visual cues of their intent, as well as current and future motion through their articulated pose~\cite{camara2020pedestrian1,camara2020pedestrian2,rasouli2019autonomous}. Human motion is predictable up to one second with around one centimeter average per joint error when observing articulated motion~\cite{lyu20223d}. The motion information present in the pedestrian pose is unused in most AV motion planning models~\cite{bae2023set,shen2018transferable,li2020learning,rhinehart2019precog,yao2021bitrap,huang2020long,deo2020trajectory,zhao2021tnt,mangalam2020not,li2022evolvehypergraph,chen2022hierarchical,djuric2020uncertainty,liang2020pnpnet,luo2022gamma,ma2019trafficpredict,zhu2020robust,sriram2020smart,zhao2019multi,luo2021safety,fang2020tpnet,park2020diverse,vansafecritic,yang2020traffic,cheng2021amenet,giuliari2021transformer,AndersonVJ20a,SalzmannICP20,HamandiDF19,Yao2019FollowingSG,ChenLSL20,IvanovicLSP21,GirgisGCWDKHP22,tang2023collaborative,huang2020diversitygan,jiang2022perceive,tang2019multiple,manish2023survey,gilles2022thomas,li2020end,zeng2020deep,casas2020spagnn,messaoud2021trajectory}, as well as in AV model testing. Progress in articulated pedestrian modeling is slowed down by the lack of data due to the difficulty in recovering articulated pedestrian poses in real traffic scenarios. The importance of preserving the relationship between pedestrian motion and scene semantics on pedestrian motion perception is shown in \Figure{problem}. The lack of data has lead to the development of AV scene understanding models~\cite{bae2023set,shen2018transferable,li2020learning,rhinehart2019precog,yao2021bitrap,huang2020long,deo2020trajectory,zhao2021tnt,mangalam2020not,li2022evolvehypergraph,chen2022hierarchical,djuric2020uncertainty,liang2020pnpnet,luo2022gamma,ma2019trafficpredict,zhu2020robust,sriram2020smart,zhao2019multi,luo2021safety,fang2020tpnet,park2020diverse,vansafecritic,yang2020traffic,cheng2021amenet,giuliari2021transformer,AndersonVJ20a,SalzmannICP20,HamandiDF19,Yao2019FollowingSG,ChenLSL20,IvanovicLSP21,GirgisGCWDKHP22,tang2023collaborative,huang2020diversitygan,jiang2022perceive,tang2019multiple,manish2023survey,gilles2022thomas,li2020end,zeng2020deep,casas2020spagnn,messaoud2021trajectory} that are oblivious to pedestrian poses and other visual cues (such as facial expressions etc), thus simply omitting available motion cues. Further AV testing is not yet utilizing realistic articulated pedestrian models and instead tests AV's interactions with heuristic pedestrian motions~\cite{zhang2022finding,zhong2021survey,wenhao2022,zhong2022,Hamdi0G20,Gupta2020TowardsSS,Mingyun20,KorenK19,MuktadirW22,ding2022causalaf,Abeysirigoonawardena19,AbdessalemPNBS18,LiTW20,AbdessalemNBS16,zhong2022neural,bussler2020application,almanee21,Wenhao21,Parashar18,Demetriou2020ADL,NishiyamaCMSHOR20,DingXZ20,Sun21,karunakaran2020efficient,KarunakaranWN20,DingCXZ20,wang2021advsim}. Since AV's are not evaluated in interactions with real humans at scale the possible safety issues in pedestrian detection, tracking and forecasting are relatively unknown. 

 We argue that articulated semantically grounded pedestrian sensing and modeling is currently an underdeveloped research field due to a lack of Ground Truth (GT) data. Supervised articulated human sensing models\cite{wei2016convolutional,cao2017realtime,cao2019open,zhang2022mixste,cheng20203d,shan2022p,chen2021anatomy,zheng20213d,LiuSW0CA20,LinL19,Zanfir_2018_CVPR,Tripathi_2023_CVPR} are often evaluated on clean benchmarks\cite{ionescu2013human3,sigal2010humaneva,mono-3dhp2017,JooLTGNMKNS15,ZanfirZBFSS21,HuangYHSAPSB22} where humans are and clearly and often fully visible, close to the camera and captured in good lightning conditions. This  leads to methods that fail at a distance as well as in the presence of motion blur or poor lightning and occlusions. Unsupervised \cite{wandt2022elepose,hu2021unsupervised,huang22occluded,XuCL021,KunduSJYJCR21,DengSZS21,KunduSJRBC20,KunduSYJCB22,Zanfir_2021_CVPR,ZanfirZBFSS21} and weakly supervised\cite{zhu2022motionbert,shan2023diffusion,zhou2019hemlets,LiLZXY22,Popa_2017_CVPR,NEURIPS2021_a1a2c3fe,ZanfirBXFSS20,GholamiWRW022,UsmanTSS22,GongZF21,RoyCHF22} training have become popular to overcome the lack of difficult and varied GT data. These models could however be improved with combined temporal and traffic-centered semantic modeling to obtain human 3D pose tracking at scale from a moving vehicle. 
 
 A ground truth dataset of articulated human motion in 3D would allow one to evaluate the discrepancy between the true and estimated scale and depth, robustness to occlusions, and motion blur in human pose detection and forecasting. In parallel to this work, an approximated dataset of articulated humans in the wild has been released\cite{von2018recovering}, but the dataset still exhibits humans that are close to the camera in the presence of little camera motion when compared to images from traffic and lacks annotations in the presence of large occlusions. Even though \cite{von2018recovering} is a step in the right direction it does not express the full complexity of the problem of articulated pedestrian motion estimation from onboard vehicles. 
 
 Existing monocular absolute scale depth estimators generalize poorly on previously unseen scenes\cite{ming2021deep,roussel2019monocular}. The same may be expected of the partially supervised and unsupervised 3D human sensing models\cite{wandt2022elepose,hu2021unsupervised,huang22occluded,XuCL021,KunduSJYJCR21,DengSZS21,KunduSJRBC20,KunduSYJCB22,Zanfir_2021_CVPR,ZanfirZBFSS21,zhu2022motionbert,shan2023diffusion,zhou2019hemlets,LiLZXY22,Popa_2017_CVPR,NEURIPS2021_a1a2c3fe,ZanfirBXFSS20,GholamiWRW022,UsmanTSS22,GongZF21,RoyCHF22}, and this is likely to also affect the estimated limb lengths of the pedestrian. Correctly estimated limb lengths however allow for a precise estimation of the pedestrian's travelling speed. Note that a moving camera requires a robust and temporally smooth pedestrian sensor and motion model to deal with possible image blur, occlusions and to avoid confusion between the motion of the pedestrian and the camera. Robust and complete pedestrian motion sensing and prediction may directly reduce the number of lethal collisions with AVs.

Pedestrian trajectory forecasting is hard because pedestrians appear to  move stochastically when compared to the more regular motion of cars, in particular when pedestrians are modelled by their bounding (3d) boxes~\cite{bae2023set,shen2018transferable,li2020learning,rhinehart2019precog,yao2021bitrap,huang2020long,deo2020trajectory,zhao2021tnt,mangalam2020not,li2022evolvehypergraph,chen2022hierarchical,djuric2020uncertainty,liang2020pnpnet,luo2022gamma,ma2019trafficpredict,zhu2020robust,sriram2020smart,zhao2019multi,luo2021safety,fang2020tpnet,park2020diverse,vansafecritic,yang2020traffic,cheng2021amenet,giuliari2021transformer,AndersonVJ20a,SalzmannICP20,HamandiDF19,Yao2019FollowingSG,ChenLSL20,IvanovicLSP21,GirgisGCWDKHP22,tang2023collaborative,huang2020diversitygan,jiang2022perceive,tang2019multiple,manish2023survey,gilles2022thomas,li2020end,zeng2020deep,casas2020spagnn,messaoud2021trajectory}. In general pedestrian motion prediction is hard as the goal of the pedestrians and the reason for a particular speed is unknown even if articulated motion is available. But pedestrians plan their motion in the scene depending on the geometry of the semantics surrounding them; for example, pedestrians may cross the road to avoid staying on a pavement that is very shallow and is next to a densely trafficked road~\cite{camara2020pedestrian2,rasouli2019autonomous}. Further, pedestrian dynamics depend on the particular pedestrian's physique~\cite{lyu20223d}. A complete pedestrian forecasting model should therefore be semantically aware as well as articulated. Currently, to our knowledge only~\cite{Priisalu_2021_CoRL} present an articulated semantically reasoning pedestrian forecasting model. A key difficulty in training articulated and semantically reasoning pedestrian models lies in the lack of data as mentioned before, but also in the lack of varied data. Pedestrians act often monotonely in traffic \cite{HugBHA21,SAADATNEJAD2022103705,SchollerALK20} as complex behaviors occur seldom in traffic, and existing datasets often do not express the full variability in human dynamics and appearances. To be utilized on-board in real time further research is necessary into robust real-time articulated semantically reasoning pedestrian motion models. 

\begin{figure}
\centering
\vspace{-0.5cm}
\includegraphics[width=\textwidth]{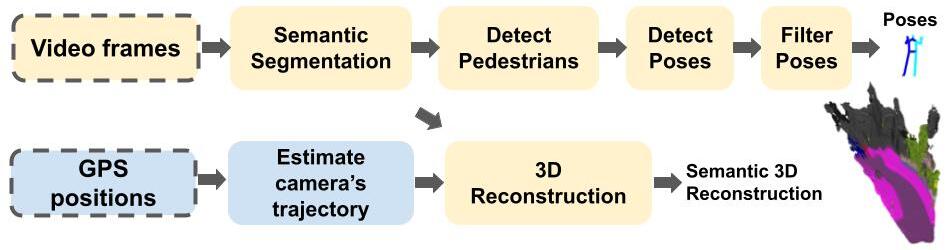}
\vspace{-1cm}
\caption{The modular reconstruction process: First the data capturing the vehicle's trajectory is estimated from GPS coordinates or accelerometer data, this is then used to initialize camera matrices in the 3D reconstruction of the scene. The frames of the binocular video are semantically segmented, semantic segmentation is used to remove moving objects (vehicles and people), and the background objects are 3D reconstructed. The semantic segmentation (or images) is then used to find the bounding boxes (BBox) of pedestrians. Then pose estimation is performed followed by filtering to disallow physically unplausible poses. Note that from semantic segmentation 3D BBox of cars can be estimated.} \label{overview}
\vspace{-0.4cm}
\end{figure}
 Autonomous vehicles typically have a number of sensors that all together generate large amounts of data (possibly up to the order of Tb per minute), so the data must be filtered for salient objects. By filtering the data we stand at the risk of possibly missing something important like a partially occluded pedestrian. Therefore how to best represent a traffic scene for autonomous driving is still an open research topic ~\cite{2018Junyaoisit,li2022bevsurvey,23signhsurround,Casas2020ImplicitLV,XiongLearningCR}. Within motion planning High Definition (HD) maps containing scene details in a compact representation ~\cite{Ma2019ExploitingSS}, and Bird's Eye View (BEV) images that is to say top view image of the scene, are common because they allow for 2D vision models to easily be utilized on traffic data\cite{li2022bevsurvey}.  Both HD and BEV are compressed scene representations that do not in general allow for sensor data augmentations.
In this work we opt for a semantically labeled 3D reconstruction with articulated pedestrians because this allows for detailed modeling of pedestrians, the evaluation of physical distances between objects, and data augmentation for a number of sensors (such as camera and LiDAR). This is an uncompressed scene representation that allows for data augmentation and testing but it is difficult to recover from only onboard binocular video, as will be detailed. Human sensing is performed on images~\cite{belkada2021pedestrians,rasouli2021bifold,agrawal2021single,huynhaol,piccoli2020fussi,ranga2020vrunet,lorenzo2020rnn,mangalam2020disentangling,kim2020real,adeli2020socially,rempe2020contact,minguez2018pedestrian,cao2020long} because this is a more mature research field than human sensing from other sensor data such as LiDAR-scans~\cite{zhao2019probabilistic,zhang2020stinet,harley2020tracking,zanfir22semi,Shah2020LiRaNetET}. 

Recovering a semantic 3D model of the scene with articulated pedestrians can be done modularly as shown in \Figure{overview} by estimating the recording device's motion, semantically segmenting the scene, and 3D reconstructing the scene. Adding articulated pedestrians to the 3D scene reconstruction requires detecting the pedestrians in the scene, estimating the pose of the pedestrians in 2D, estimating the pose of the pedestrians in 3D, and filtering any physically unrealistic poses. A number of estimation errors can occur along the way, making such data gathering hard. We hypothesize that articulated human sensing, tracking and prediction could be improved by combining the three tasks, as is done for vehicles in \cite{agro2023implicito,Li2020EndtoendCP,Shah2020LiRaNetET,Sadat2020PerceivePA,Liang2020PnPNetEP,Sadat2019JointlyLB}. After the development of the presented results pose tracking has been posed as the problem of tracking the pose of one or more pedestrians \cite{xu2020hi,chang2020towards,li2020simple}. 

Note that even though human motion can be captured with a Motion Capture (MoCap) system, or recently even from selected images~\cite{von2018recovering}, it is not trivial to set up large-scale experiments to gather traffic datasets that contain a large variety of possible scene geometries, semantics, and GT poses. This is because MoCap data gathering requires intervening with the scene, and existing human pose sensing methods from images cannot yet capture the poses of all humans in the images ~\cite{von2018recovering}. Further, most MoCap methods cannot be utilized accurately outdoors with large occlusions. More research is needed in human motion capture in traffic. Markerless human pose detection results often look impressive~\cite{saini2023smartmocap}, but don't often present any results for humans who are far away in the presence of motion blur, which is the case in traffic data. Human detection at a distance in the presence of motion blur is still challenging, let alone human pose detection. Other sensors can be used to remedy motion blur and aid in human detection\cite{zhang_cvpr2023_oyster} and articulated human sensing, for example \cite{windbacher3dpose} perform an initial step to utilize LiDAR and images to detect distant humans in real traffic data.

\begin{figure}
\centering
\includegraphics[width=\textwidth]{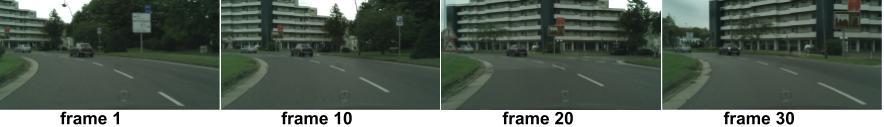}
 \vspace{-0.9cm}
\caption{A sub-sampled sequence of frames from the Cityscapes dataset, Aachen.}\label{fig:frames}
\end{figure}
\section{Scene reconstruction}
We use the Cityscapes dataset\cite{Cordts2016Cityscapes} that consists of binocular video sequences, with a length of 30 frames at 17 frames per second, gathered from calibrated cameras placed on the front screen of a vehicle. Sample images are shown in \Figure{fig:frames}. The data gathering vehicle's position can be estimated from the provided GPS coordinates or accelerometer data. Disparity maps are provided for each frame, and a GT semantic segmentation is provided of the leftmost camera's image at the 20th frame. The images contain some blur because they are captured from behind the windscreen as the vehicle moves. Image blur, the fast camera motion in the forward direction (most 3D reconstruction methods are fragile to this) and independently moving objects make 3D reconstruction of the sequences hard. The inherent difficulties in 3D reconstructing onboard videos have led to the increased popularity of LiDAR for depth estimation.
\begin{figure}[h!]
\centering
  \includegraphics[width=\textwidth]{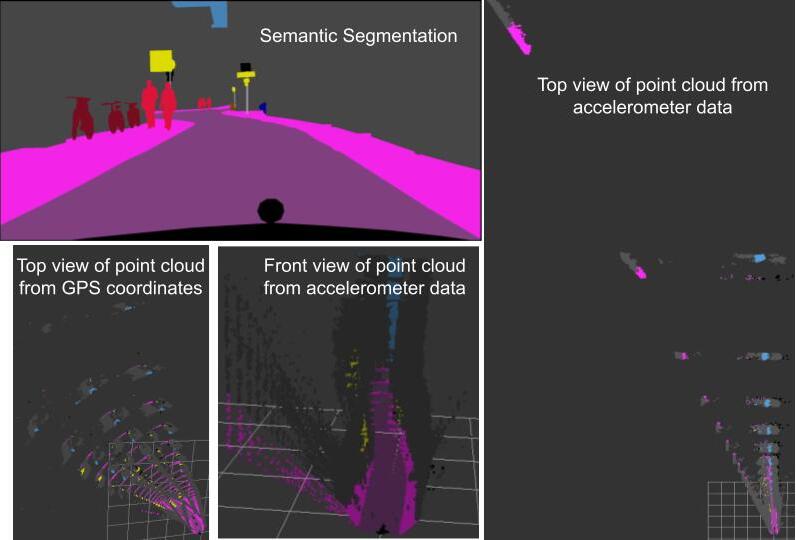}
   \vspace{-0.8cm}
\caption{Visualizations of 3D pointclouds from using the vehicle's GPS coordinates or accelerometer readings to estimate camera position with per-frame disparity maps. The GPS is noisier than the accelerometer resulting in a noisier pointcloud.}\label{camPos}
\end{figure}
\subsection{ Initial Camera Positions}
Assuming that the cameras cannot move within the rig or the car we can estimate the cameras' motion as the vehicle's motion. The vehicle's motion can be estimated from the Global Positioning System (GPS) or the accelerometer data. The GPS coordinates contain jumps as seen in \Figure{camPos} where the cameras' estimated position and each frame's disparity map are used to create pointclouds for each frame that are then aggregated. It can be seen that in the GPS-based vehicle trajectory, the vehicle's rotation oscillated from frame to frame causing the 3d point clouds of different frames to diverge, while the accelerometer data results in a smoother pointcloud. This suggests that the accelerometer-based vehicle trajectory is a better initialization for the camera matrices in a 3D reconstruction system.
\subsection{ 3D Reconstruction}
\begin{table}[p]
\caption{Overview of the algorithms used by the different SFM and SLAM libraries. See the Appendix for more details.}
 \vspace{-0.3cm}
\begin{tabular}{llllll}
\toprule
Method     & \begin{tabular}[c]{@{}l@{}}Image \\ features\end{tabular} & \begin{tabular}[c]{@{}l@{}}Matching \\ algorithm\end{tabular}             & \begin{tabular}[c]{@{}l@{}}First view \\ selection\end{tabular}                    & \begin{tabular}[c]{@{}l@{}}Method for \\ selecting addi-\\ tional views\end{tabular} & \begin{tabular}[c]{@{}l@{}}Bundle \\ Adjustment\end{tabular}   \\ \midrule
OpenSFM    & HAFP+HOG                                                  & \begin{tabular}[c]{@{}l@{}}Exhaustive\\ Fast approx. NN\end{tabular}      & \begin{tabular}[c]{@{}l@{}}First frames\\ \textgreater{}30\% outliers\end{tabular} & \begin{tabular}[c]{@{}l@{}}largest overlap\\ with pointcloud\end{tabular}            &                                                                \\
Bundler    & SIFT                                                      & \begin{tabular}[c]{@{}l@{}}Exhaustive\\ approx NN\end{tabular}            & \begin{tabular}[c]{@{}l@{}}large  difference\\ in rotation\end{tabular}            & \begin{tabular}[c]{@{}l@{}}largest overlap \\ with pointcloud\end{tabular}           & SPA                                                            \\
OpenCv  & DAISY                                                     & Exhaustive NN                                                             &                                                                                    &                                                                                      & \begin{tabular}[c]{@{}l@{}}inexact \\ Newton\end{tabular}      \\
VisualSFM  & SIFT GPU                                                  & \begin{tabular}[c]{@{}l@{}}Preemptive \\ mathcing\end{tabular}            & \begin{tabular}[c]{@{}l@{}}thresholded no. \\ of large features\end{tabular}       & \begin{tabular}[c]{@{}l@{}}largest overlap\\ with pointcloud\end{tabular}            & \begin{tabular}[c]{@{}l@{}}Multicore \\ BA\end{tabular}        \\
ORBSLAM    & ORB                                                       & \begin{tabular}[c]{@{}l@{}}Stereo matching\\ closer than 40b\end{tabular} & first frames                                                                       & next frame                                                                           & \begin{tabular}[c]{@{}l@{}}Levenberg \\ Marquardt\end{tabular} \\
COLMAP     & SIFT                                                      & Exhaustive NN                                                             & \begin{tabular}[c]{@{}l@{}}Algorithm of  \\ Beder \& Steffen\end{tabular}          & high inlier ratio                                                                    & PCG                              \\ \bottomrule                            
\end{tabular}\label{table:3d_reconstruction_models}
\end{table}
\begin{figure}[p]
\centering
\includegraphics[width=.7\textwidth]{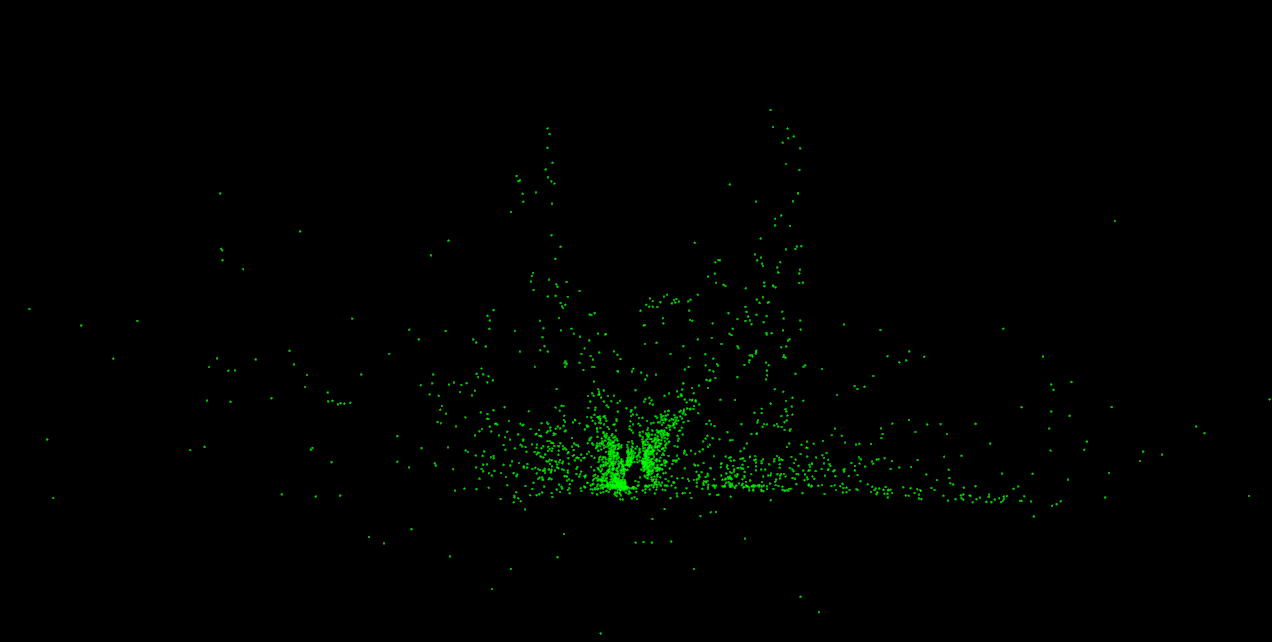}
 \vspace{-0.4cm}
\caption{3D pointcloud reconstructed by the OpenCV library. There are too few 3D points in the pointcloud to detect what has been reconstructed.}\label{fig:opencvBA}
\end{figure} %% Add Ordblsam results
\begin{figure}[p]
\centering
  \includegraphics[width=\textwidth]{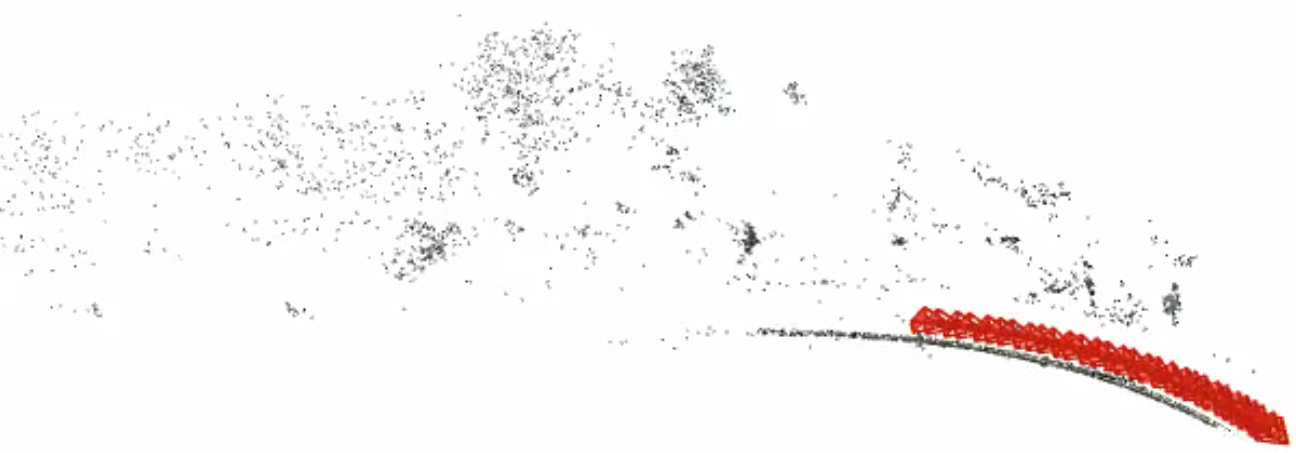}
  \vspace{-0.8cm}
\caption{COLMAP's sparse 3D reconstruction of the scene depicted in \Figure{fig:points} and \Figure{camPos}. The red rectangular pyramids depict the different camera positions, showing correctly that the vehicle traveled on a curved road.}\label{fig:sparse}
\end{figure}
Multiple 3D reconstruction methods were tested, but only COLMAP\cite{morelli2023colmap} converged on a large number of the available sequences. It should be noted that all libraries were tested on the same three sequences, all containing some moving pedestrians and vehicles and strong forward motion as this is typical for the Cityscapes dataset. The following libraries were tested with the following results:
% \begin{itemize}
%     \item Open Structure for Motion Library (OpenSFM)\cite{Mapillary}- Fails to reconstruct the Cityscapes scenes.
%      \item Bundler\cite{snavely2006photo}- Fails to reconstruct the Cityscapes scenes.
%      \item OpenCV SFM library\cite{opencv} - Reconstruction with unclear structure. See \Figure{fig:opencvBA}
%      \item VisualSFM\cite{wu2013towards}-Fails to reconstruct the Cityscapes scenes.
%      \item ORBSLAM\cite{murartla2017orb} -  Reconstruction with unclear structure.
%      \item COLMAP\cite{schoenberger2016sfm,schoenberger2016mvs} - Reconstructs 150 of the 3475 scenes available. See \Figure{fig:sparse}.
% \end{itemize}
 \emph{Open Structure for Motion Library} (OpenSFM)\cite{Mapillary}- A Structure from Motion(SFM) system, that is an incremental 3d reconstruction system. Fails to reconstruct the Cityscapes scenes likely because the change in camera rotation is too small between frames. Bundler\cite{snavely2006photo} is also a SFM system. Finds <10 matches, and fails again likely because the images are blurry and the rotational difference between the initial camera views is too small. \emph{OpenCV Structure from motion library}\cite{opencv} - A SFM library that uses DAISY features\cite{tola2009daisy}. Result of 30 frames - finds relatively few points without a clear structure. See \Figure{fig:opencvBA}.   \emph{VisualSFM}\cite{wu2013towards} a paralellized SFM pipeline with Bundler. Only a thresholded number of large-scale features are matched across images. This unfortunately fails possibly because of image blur or the lack of distinct large-scale structures in the images. The method is unable to find enough SIFT feature points likely because the images are blurry and finds no verified matches between two stereo images. Finally, VisualSFM cannot handle forward motion, not finding a good initial pair of images with enough matches.  \emph{ORBSLAM}\cite{murartla2017orb}- ORB-feature\cite{rublee2011orb} (a fast feature descriptor combining gradient and binary features) based Simultaneous Localization And Mapping (SLAM) system. Finds too few keypoints, likely due to blur and depth threshold. Results in a too sparse reconstruction.
 \emph{COLMAP}\cite{schoenberger2016sfm,schoenberger2016mvs}- an incremental SFM and Muti-view stereo(MVS) system. Extracts SIFT\cite{lowe2004sift} features that are exhaustively matched (other matching methods are also available) across all images. Converges for 150 scenes on the training and validation set and 150 scenes on the test set. See \Figure{fig:sparse}. For further details on the different systems see \Table{table:3d_reconstruction_models} and the Appendix.

A number of the reconstruction methods fail to find reliable matches across images, likely because of the motion blur and poor quality of the images as the cameras are mounted behind the windscreen of the vehicle. Secondly, the majority of visual 3D reconstruction methods fail at reconstructing in the presence of large forward motion of the camera in particular at fast speeds (i.e. the speed of a car) in the presence of a large number of objects at a large distance to the camera.  
COLMAP \cite{schoenberger2016sfm,schoenberger2016mvs} differs mostly from the other methods by the fact that it is modeled for camera views with at times large overlaps; by its outlier robust triangulation, probabilistic new view selection and iteratively applied final Bundle adjustment (BA) alternated with filtering and triangulation. It should, however, be noted that COLMAP is not applicable in real-time, (recently a real-time adaption\cite{morelli2023colmap} has become available), and it could not reconstruct all of the Cityscapes scenes ( 3475 in the training set and 1525 in the test set). The majority of 3D reconstruction methods are not well-fitted to reconstruct images captured from a moving vehicle. This has led to the increased popularity of LiDAR sensors as they can measure directly the distance to objects which is particularly useful in the presence of moving objects (pedestrians, cars, bikers etc.) when only a few images may be available of the object at a particular location. 

\subsection{Filtering out Non-stationary Objects in the 3D Reconstructions}
\begin{figure}[p]
\centering
         \includegraphics[width=\textwidth]{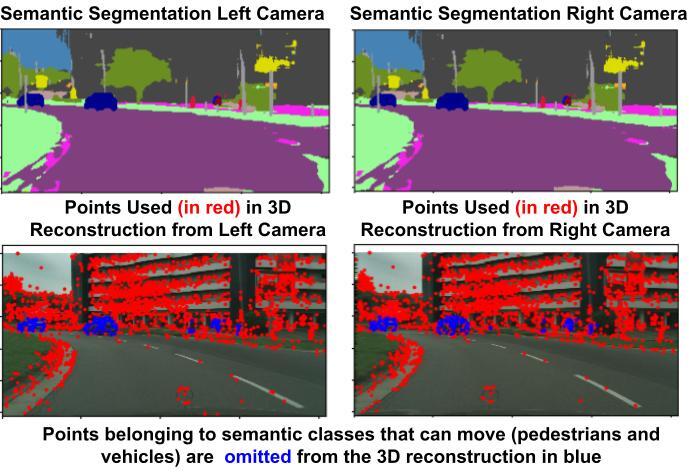} \vspace{-1cm}
         \caption{Points belonging to moving objects cannot be used in the SFM and must filtered out. In the top row, the semantic segmentation of the left and the right camera images are shown for one frame, and in the bottom row, the points used in the sparse reconstruction of COLMAP are shown. In red points that are included in the SFM are shown. In blue points that are omitted in the SFM (as they belong to semantic classes of pedestrians and vehicles) are shown.}\label{fig:points}
\end{figure}
\begin{figure}[p]
\centering
 \includegraphics[width=\textwidth]{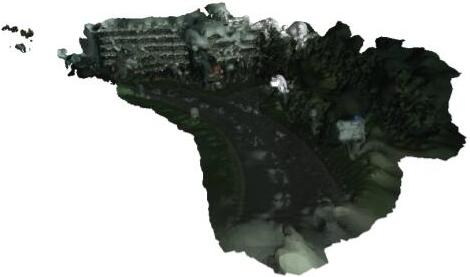}
 \vspace{-1cm}
\caption{COLMAP's dense 3D reconstruction of the scene depicted in \Figure{fig:points} and \Figure{camPos}. The dense reconstruction is noisy but the scene is recognizeable.}\label{fig:dense}
\end{figure}

Moving objects such as cars and pedestrians need to be removed in SFM, to this end the Gated Recurrent Flow Propagation (GRFP) net\cite{nilsson2018semantic} that is a video segmentation network that utilizes optical flow to stabilize semantic segmentation in video data. The GRFP is used to segment the frames of the Cityscapes sequences.

 COLMAP is adapted by adding semantic segmentation as an additional channel (in addition to the 3 RGB channels) describing points during SFM. The camera matrices are initialized based on the accelerometer data. A subsampled sequence of frames from the left camera can be seen in \Figure{fig:frames} and the semantic segmentation of the last frame in the left and the right images can be seen in \Figure{fig:points}. The resulting sparse reconstruction in \Figure{fig:sparse} and dense reconstruction can be seen in \Figure{fig:dense}. Semantic segmentation of the reconstructed 2D points is then transferred to the 3D pointcloud as detailed in the supplementary material of \cite{Priisalu_2020_ACCV}. Note that moving objects are filtered out only during sparse reconstruction, the dense reconstruction is instead filtered for objects with dynamic object labels during voxelisation. 
%------
\begin{figure}[p!]
\centering
         \includegraphics[width=\textwidth]{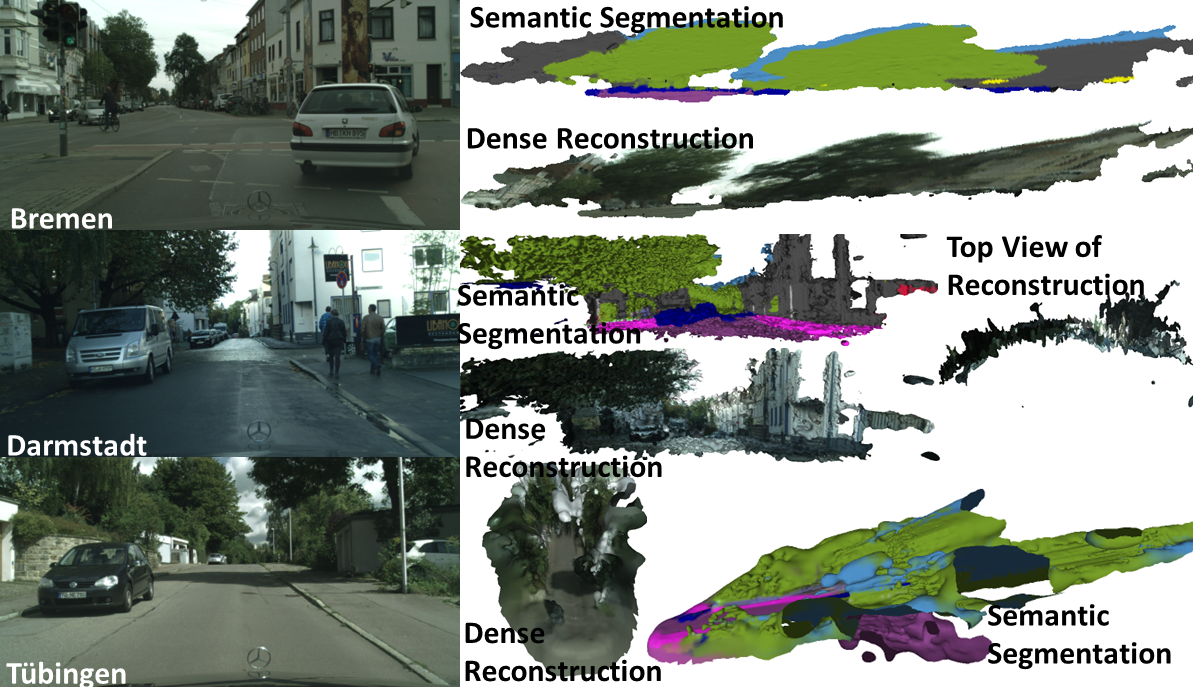}
          \includegraphics[width=\textwidth]{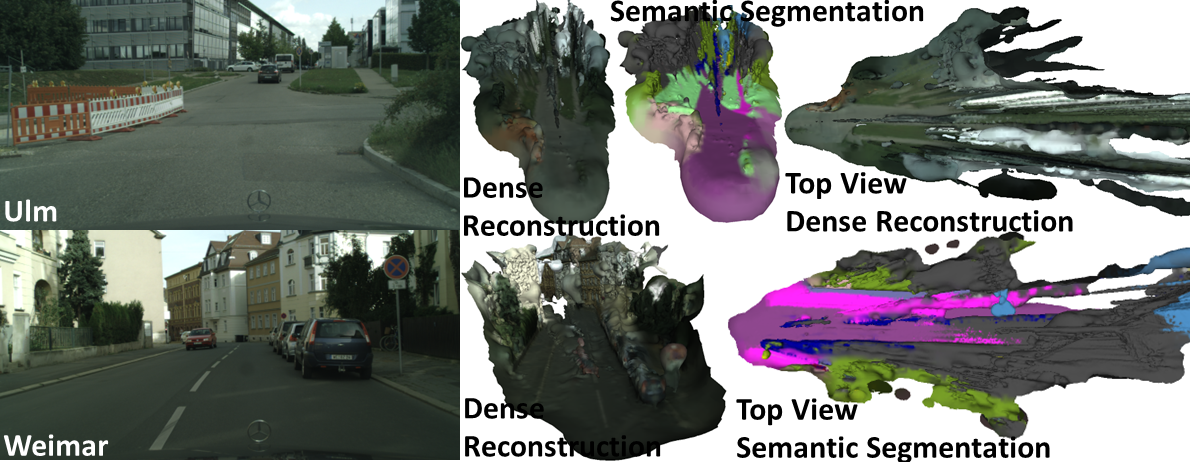}
           \vspace{-0.8cm}
         \caption{Dense reconstructions. \emph{First row:} The Bremen sequence's first frame (to the left) and a flat reconstruction (to the right) labeled with semantic segmentation labels (top) and RGB (bottom).\emph{Second row:} The Darmstadt's reconstruction appears fine from the front view (middle) but is flat and curved when viewed from the top (left).\emph{Third row:} Tübingen results in a correct reconstruction of the street close to the camera (middle), but an incorrect estimation of the street topology due to uphill view (right).\emph{Fourth row:} Ulm is reconstructed correctly with a patch of grass separating the road and the sidewalk as seen in front (middle) and top view(to the right).\emph{Fifth row:} Correctly reconstructed street shape as seen in front (middle) and top view (to the right).}
         \label{fig:multiple_reconstructions}
\end{figure}

\begin{figure}[h!]
\centering
         \includegraphics[width=\textwidth]{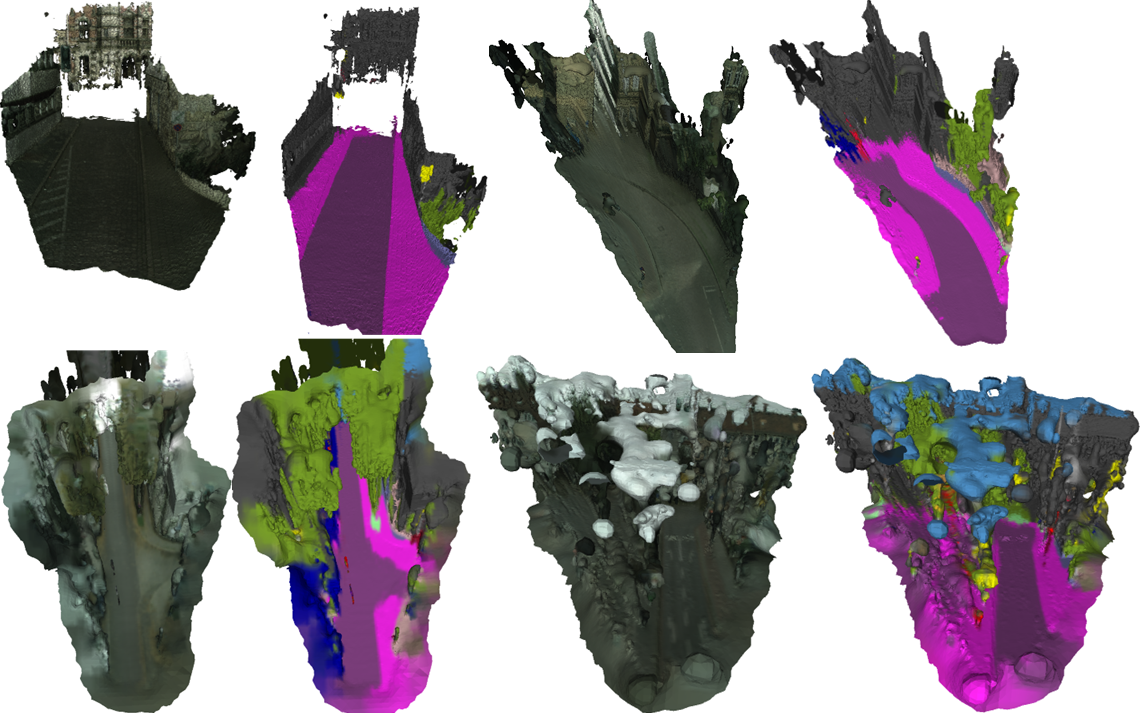}
         \vspace{-0.8cm}
         \caption{Additional dense reconstructions from Bremen showing that noise levels vary but the street shape is often successfully reconstructed.}
         \label{fig:additional_results}
\end{figure}

A number of reconstructions are shown in \Figure{fig:multiple_reconstructions}.
Some reconstructions correctly recover the structure of the road such as Tübingen, Ulm and Weimar, also in \Figure{fig:additional_results}. In general, the reconstruction deteriorates further away from the camera. This can be seen in the reconstruction of Tübingen in \Figure{fig:multiple_reconstructions} where some of the road (in purple) is misaligned with the rest of the reconstruction and is tilted downwards. This is expected as objects further away from the camera are harder to recognize and estimate the distance to. The reconstructions elongate objects as can be seen in the reconstruction of Tübingen Ulm and Weimar in \Figure{fig:multiple_reconstructions}. COLMAP is however not always successful, when the frames change in viewpoints is small the found reconstruction ends up being flat like in Bremen in \Figure{fig:multiple_reconstructions} or almost flat like seen in the top view of Darmstadt in \Figure{fig:multiple_reconstructions}. 
%------
\begin{table}[h]
\centering
\caption{Frequency of the semantic classes on the CARLA dataset and accuracy of the \emph{Pointnet++} for the different semantic classes. The semantic classes are in the order of decreasing frequency. Objects of class wall obtains the lowest accuracy.}
 \vspace{-0.3cm}
\begin{tabular}{l|c|c|c|c|c|c|c|c|c}
\toprule  
\bf{Class} &\bf{vegetation}&\bf{building}&\bf{road}&\bf{sidewalk}&\bf{fence}& \bf{static}& \bf{wall}& \bf{pole}& \bf{sign}\\\midrule
 \bf{Frequency} $\%$ &37.62&18.50&17.87&17.55&3.00&2.81&1.71&0.83&0.09\\
 \bf{Accuracy}&0.93& 0.86&0.88&0.67&0.84&0.51&\textbf{0.50}&0.88&0.71\\\bottomrule
\end{tabular}\label{tab:poitnet}
\end{table}

To directly label a poinctloud experiments were conducted with the popular pontcloud segmentation network \emph{Pointnet++}\cite{qi2017pointnet++}. The Cityscapes has no GT segmented pointclouds, so a model that was finetuned on CARLA\cite{dosovitskiy2017carla} generated pointclouds was tested but resulted in confused labels. Finetuning of Pointnet++\cite{qi2017pointnet++} on the synthetic CARLA dataset(from \cite{Priisalu_2020_ACCV}) resulted in a low mean average class accuracy of 0.62, with per class results shown in \Table{tab:poitnet}. The classes that occur seldom get low accuracy, so objects such as traffic sign get almost always incorrectly labelled. It is also worth noting that strangely enough points belonging to walls are correctly marked only in half of the occurences. In general the results suggest that Pointnet++ results are not on bar with labelling 3D reconstructions according to projections of 2D semantic maps. It is possible that more recent methods\cite{schult2023mask3d,ngo2023isbnet,sun2023superpoint,chen2023clip2scene,cheng20212,hou2022point,xu2021rpvnet,zhu2021cylindrical} could improve the results.

\begin{figure}[h]
\centering
\includegraphics[width=0.95\textwidth]{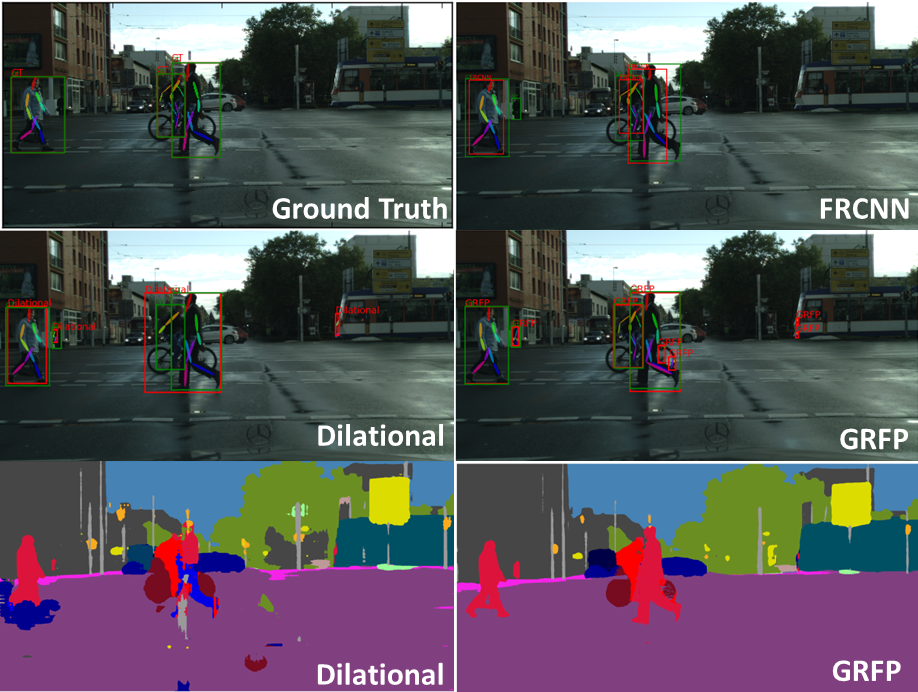}
\vspace{-0.5cm}
\caption{Segmentation, BBoxes and 2D joint position estimates of \emph{OpenPose} with \emph{Dilational}, \emph{GRFP} and \emph{FRCNN}. \emph{Dilational} net and \emph{GRFP} manage to separate different pedestrians who are visually close by but also introduce false positives. \emph{GRFP} produces cleaner BBoxes than \emph{Dilational}.}\label{fig:seg}
\end{figure}
\section{Pedestrian sensing}
% Pedestrian detection
% Detecting 2D poses
% Filtering poses.
Detecting humans is hard because they are relatively small in traffic images, they vary in physique and visual qualities depending on the human's pose and clothing, and they change their positions from frame to frame. The fact that most popular object detectors are biased to detect close-up objects centered in an image makes them ill-fitted to traffic data because in traffic humans appear often a distance from the camera. We compared object detection, segmentation, and human pose detecting network's ability to detect pedestrians on the Cityscapes dataset\cite{Cordts2016Cityscapes} by comparing the detected pedestrian's BBox overlap with BBoxes generated from GT segmentations. The tested methods are
\begin{itemize}
    \item \emph{DilationalNet-10} \cite{yu2015multi}- A popular semantic segmentation network with dilated convolutions for larger receptive field.
    \item \emph{The Gated Recurrent Flow Propagation(GRFP)\cite{nilsson2018semantic}} - A temporally smoothed video segmentation network showing temporally smooth result on the Cityscapes dataset\cite{Cordts2016Cityscapes}.
    \item \emph{Faster-RCNN(FRCNN)}\cite{ren2015faster} - A popular object detection network with high throughput and good performance on benhcmarks.
    \item \emph{OpenPose}\cite{cao2017realtime,cao2019open} - A popular multi human 2D pose estimating network, that has a runtime that scales well with increasing number of visible humans.
\end{itemize}

\begin{figure}[h]
\centering
 \includegraphics[width=\textwidth]{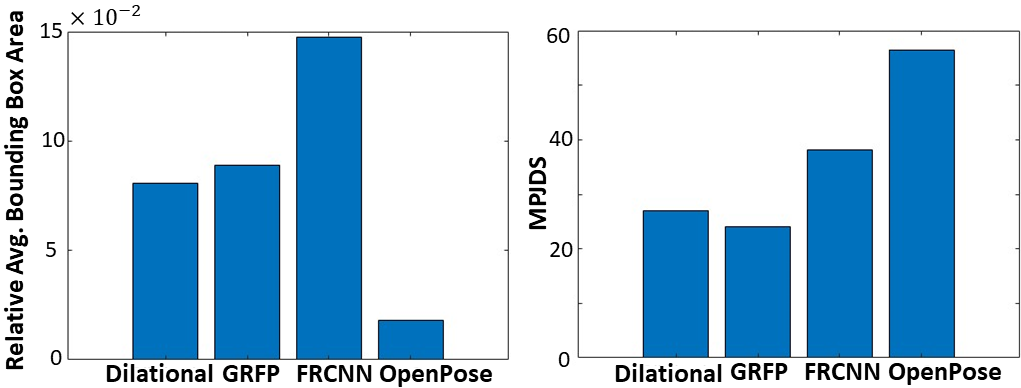}
 \vspace{-0.8cm}
\caption{\emph{Left:} The average relative BBox areas of the different pedestrian detection methods. FRCNN detects on average the largest BBoxes and OpenPose the smallest. \emph{Right:} The average distance from estimated joint positions to human mask (from GT segmentation). GRFP's human BBoxes result in the lowest distance from estimated joint positions to human mask and OpenPose in the largest.}\label{fig:pedestrian_bbox}
\end{figure}

\begin{table}[h]
\centering

\caption{The number of true positive, false negative and false positive BBoxes on the training and validation sets of Cityscapes for the 20th frame. The two strongest contenders for pedestrian detection are the \emph{GRFP} and the \emph{FRCNN}. The \emph{GRFP} produces the largest number and area of true positives, and the \emph{FRCNN} produces the smallest number and area of false positives and negatives.}
\vspace{-0.2cm}
\begin{tabular}{lcccccc}
\toprule  
Model &\textbf{True} & \textbf{Average }&\textbf{False} & \textbf{Average}&\textbf{False}& \textbf{Average}\\ 
&\textbf{positives}&\textbf{TP area}&\textbf{positives}& \textbf{FP area}&\textbf{Negatives}& \textbf{FP area}  \\\midrule
\textbf{Dilational}&2887 &0.699& 7,516 & 0.008 & 16,664 & 0.016 \\
\textbf{GRFP} &\textbf{3588}& \textbf{0.707}&14,317 & 0.01 & \textbf{15,980} & 0.015\\
\textbf{FRCNN} &2952& 0.706&578 & \textbf{0.001} & 16,522 & \textbf{0.009} \\
\textbf{OpenPose} &165&0.682&\textbf{343} & 0.003 & 18,997 & 0.017 \\
 \bottomrule
\end{tabular}\label{tab:bboxfalse}
\end{table}
In \Table{tab:bboxfalse} it can be seen that \emph{FRCNN} produced the smallest false positive(FP) and false negative(FN) average BBox area, but has the second highest true positive(TP) Intersection over Union (IoU) area. Because the areas of the BBoxes vary we present both the FP, FN, and TP counts and areas (normalized with respect to the total 
GT BBox areas), to observe how many individuals are detected versus how much of the visual area is covered by the pedestrians.  \emph{FRCNN} is accurate in detecting large BBoxes, and it detects on average larger BBoxes than the GRFP as seen in \Figure{fig:pedestrian_bbox} \emph{Left}. \emph{GRFP} on the other hand is better at capturing distant pedestrians but also produces a large amount of FPs. Based on this \emph{FRCNN} is the most suitable pedestrian detector as it is the most accurate in detecting pedestrians close to the vehicle, these pedestrians have the highest risk of being run over if undetected. 

 \emph{OpenPose} is used to estimate the articulated human 2D pose on the BBoxes found by \emph{Dilational}, \emph{GRFP} and \emph{FRCNN} and on the whole image.  We introduce the Mean Per Joint Distance to Segmentation(MPJDS) metric which is the average distance from an estimated 2D joint position to a pedestrian or biker segmentation mask. The MPJDS is an approximate measure of how accurately \emph{OpenPose} can estimate the pose of a pedestrian present in the BBoxes found by the different models, results are shown in \Figure{fig:pedestrian_bbox} \emph{Right}.\emph{ GRFP} results is the smallest error likely because \emph{GRFP} detects smaller BBoxes than \emph{FRCNN} resulting in smaller absolute errors. \emph{OpenPose} applied on the whole image detects pedestrians that appear to be far away from the camera, but fails to estimate their pose, resulting in large joint errors for small BBoxes. Eventhough \emph{OpenPose} has presented impressive results it fails to detect multiple pedestrians in traffic scenarios without a separate pedestrian detector. 
\begin{table}

\caption{The FRCNN detects fewer pedestrians and bikers than Dilational but results in a lower MPJDS, suggesting that FRCNN detects pedestrians that are clearer.}
 \vspace{-0.3cm}
\begin{tabular}{lcccccc}
\toprule  
\textbf{Model} & \textbf{MPJDS} & \textbf{MPJDS}& \textbf{Number of} & \textbf{Number of}& \textbf{Crossover}&\textbf{Crossover}\\ 
&& \textbf{norm.} &\textbf{pedestrians}& \textbf{bikers}&\textbf{pedestrians}& \textbf{bikers}\\
\midrule
\textbf{GT} & 32.50& 0.99&\textbf{1,803}&\textbf{17,415}& \textbf{1.0}&\textbf{1.0}\\

\textbf{Dilational}&27.17&0.87&850&4,572&0.94&0.91\\

\textbf{FRCNN} &\textbf{10.74}&\textbf{0.38}&392&2,888&0.86&0.85\\

 \bottomrule
\end{tabular}\label{tab:bbox_filtered_error}
\vspace{-0.5cm}
%\vspace{-1.5cm}
\end{table}
  \begin{figure}[h]
\centering
\includegraphics[width=\textwidth]{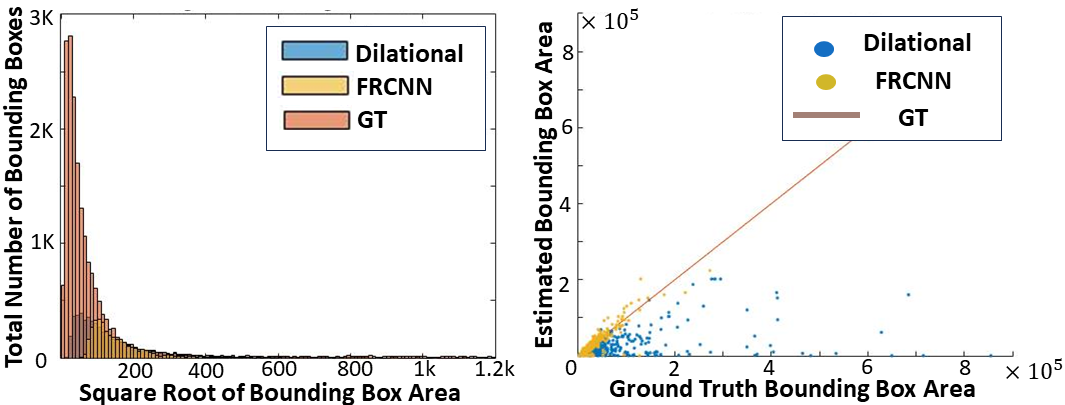}
  \vspace{-0.8cm}
\caption{\emph{Left:} \emph{FRCNN} detects only large BBoxes. \emph{Dilational} can detect smaller BBoxes, but many GT bbes are undetected by both methods. \emph{Right:} The BBoxes found by \emph{FRCNN} are in general representative of the GT BBox sizes, while \emph{Dilational} underestimates BBox sizes.}\label{fig:pedestrian_bbox_area_different_methods}
\end{figure} 

\begin{figure}[h]
\centering
\includegraphics[width=\textwidth]{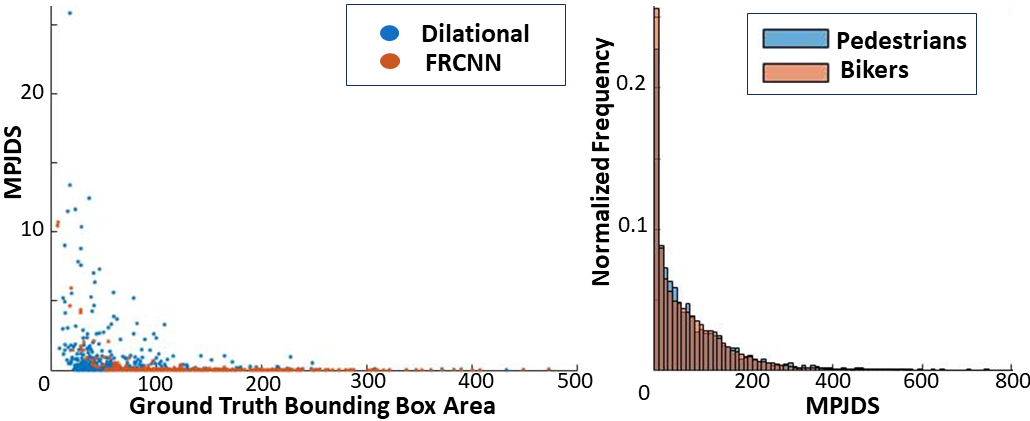}
\vspace{-0.8cm}
\caption{\emph{Left:} The MPJDS is plotted against the BBox area for the BBoxes found by the different methods. \emph{FRCNN} finds larger BBoxes and results in lower MPJDS for these larger BBoxes than for the BBoxes found by \emph{Dilational} net, suggesting that FRCNN finds easier to detect pedestrians. \emph{Right:} Histogram of MPJDS distribution of the \emph{FRCNN} detections shows that most errors are small. There appear to be no outliers with large MPJDS error. }\label{fig:pedestrian_bbox_errors_to_area}
\end{figure}

To study the accuracy of \emph{OpenPose} on BBoxes that truly contain a pedestrian we filter out the BBoxes that have at least 50\% cross-over 
 with the GT BBoxes, results are shown in \Table{tab:bbox_filtered_error}. By \emph{cross-over} is meant the percentage that the GT BBox intercepts the estimated BBox with. If an estimated BBox intercepts with a number of GT BBoxes then only the highest cross-over is recorded.
The MPJDS of \emph{OpenPose} applied on the GT BBoxes in \Table{tab:bbox_filtered_error} is much larger than that of the other two methods because the GT contains pedestrians who are hard to spot in the images (distant or occluded as seen in sizes in \Figure {fig:pedestrian_bbox_errors_to_area}). These pedestrians go unnoticed by the \emph{Dilational} and \emph{FRCNN}. Even though \emph{FRCNN} has a lower cross-over percentage than the \emph{Dilational} it obtains the lowest MPJDS suggesting that \emph{FRCNN} detects the most clearly visible pedestrians. 
 In \Table{tab:bbox_filtered_error} it can be seen that \emph{FRCNN} detects fewer pedestrians and bikers \emph{Dilational}, but results in a much lower MPJDS. The MPJDS of \emph{FRCNN} is high even though the cross-over is lower than for the other models. This is likely because \emph{FRCNN} finds pedestrians that are closer to the camera and thus clearer, omitting smaller pedestrians that are captured by \emph{Dilational} as seen in \Figure{fig:pedestrian_bbox_area_different_methods}\emph{Left}. 
 
 The \emph{FRCNN} correctly estimates the GT BBox sizes as seen in \Figure{fig:pedestrian_bbox_area_different_methods} \emph{Right}, but \emph{Dilational} underestimates BBox sizes showing the pedestrians only partially and therefore has a higher MPJDS than \emph{FRCNN} even for large GT BBoxes as seen in \Figure{fig:pedestrian_bbox_errors_to_area}. \emph{Dilational} net can detect smaller pedestrians because it has been trained on the Cityscapes dataset in difference to \emph{FRCNN}. It is possible that \emph{FRCNN} cannot detect small pedestrians because it has been trained with larger anchor sizes than the visible pedestrians.
An example showing close up occluded pedestrians comparing the \emph{GRFP}, \emph{Dilational} net, \emph{FRCNN} and GT can be seen in \Figure{fig:seg}. 
%  \begin{figure}[h]
% \centering
% \label{fig:pedestrian_bbox_distr}
%  \includegraphics[width=0.45\textwidth]{figures/pedestrian_errors/bikers.jpg}
%    \includegraphics[width=0.45\textwidth]{figures/pedestrian_errors/pedestrians.jpg}
% \caption{}
% \end{figure}

% \begin{table}
% \label{tab:bbox_filtered_size}
% \caption{  }
% \begin{tabular}{l|cccc}
% \toprule  
% Model&Mean ground truth & Mean ground truth & Mean estimated& Mean estimated \\ 
% & bbox width& bbox height& bbox width& bbox height\\
% \midrule
% ground truth &57 &62&95&152\\

% Dilational&51&61&137&166\\

% FRCNN &\textbf{85}&\textbf{98}&96&238\\

%  \bottomrule
% \end{tabular}
% \end{table}
\begin{table}

\caption{By introducing smallest size constraints on the BBoxes the number of false positives can be reduced significantly.}
\vspace{-0.3cm}
\centering
\begin{tabular}{l|cccc}
\toprule  
\textbf{Model}&\textbf{ GT}&\textbf{ GT Filtered}&\textbf{Dilational}&\textbf{GRFP}  \\ 
\midrule
\textbf{True Positives}&\textbf{15,934}&8,986&3,316&3,643\\
\textbf{False Positives}&0&0&\textbf{235}&1,038\\
 \bottomrule
\end{tabular}\label{tab:bbox_size_filtered_size}
\vspace{-0.5cm}
\end{table}

 The \emph{Dilational} net has trouble differentiating between the labels: “pedestrian”, “biker”, and “bike” as seen in \Figure{fig:seg}. Therefore \emph{Dilational} net BBoxes are fitted with skeletons after allowing biker and pedestrian labels to be interpreted as the same label. Also, bike labels are allowed to be interpreted as human if they are in connection to rider or pedestrian labels. In \Figure{fig:seg} it can also be seen that only a small change in the placement of the BBox around a pedestrian results in variations in the estimate of the pose, showing that \emph{OpenPose} is not robust to errors in pedestrian BBox placement. 
 
 Further on crowded images the \emph{FRCNN} has superior performance because it can separate between pedestrians as seen in \Figure{fig:seg3}, and \emph{GRFP} is superior in distant pedestrian detection as seen in \Figure{fig:seg2}. In the crowded scene the pose estimator gets confused with BBoxes because for some pedestrians only a single body part is visible, and it is hard for the pose estimator to detect that the body part is not just a blurry image of a human.  Videos of sample pose estimations can be found at \url{https://youtu.be/qpxpdtHbbGA} where it can be seen that the pose estimations are not temporally smooth for any of the proposed methods.
To avoid false detections of the \emph{Dilational} net and \emph{GRFP} we remove any BBoxes that are smaller than 7 pixels in width and 25 pixels in height. This results in a decrease in the number of false positives for \emph{GRFP} and the \emph{Dilational} as seen in \Table{tab:bbox_size_filtered_size}.
% \begin{table}
% \label{tab:bbox_size_filtered_size}
% \caption{Bu introducing smallest size constraints on the bounding boxes the number of false positives can be reduced significantly.}
% \begin{tabular}{l|cc}
% \toprule  
% Model& Number of pedestrians&  False positives \\ 
% \midrule
% ground truth & \textbf{15,934} &0\\%13,782&2,152
% filtered ground truth &8,986& 0\\%7,459&1,527\\
% Dilational&3,316&\textbf{235}\\
% GRFP &3,643&1,038\\
%  \bottomrule
% \end{tabular}
% \end{table}
% \begin{table}
% \caption{By introducing smallest size constraints on the BBoxes the number of false positives can be reduced significantly.}
% \centering
% \begin{tabular}{l|ccc|ccc}
% \toprule  
% \textbf{Model}&\multicolumn{3}{c}{All Bounding Boxes} &\multicolumn{3}{c}{Filtered}\\ 
% &\textbf{ GT}&\textbf{Dilational}&\textbf{GRFP}&\textbf{ GT}&\textbf{Dilational}&\textbf{GRFP}  \\ 
% \midrule
% \textbf{True Positives}&\textbf{15,934}& 2,887&3,588&8,986&3,316&3,643\\
% \textbf{False Positives}&0&7,516&14,317&0&\textbf{235}&1,038\\
%  \bottomrule
% \end{tabular}\label{tab:bbox_size_filtered_size}
% \end{table}

\begin{figure}[h]
\centering
\includegraphics[width=\textwidth]{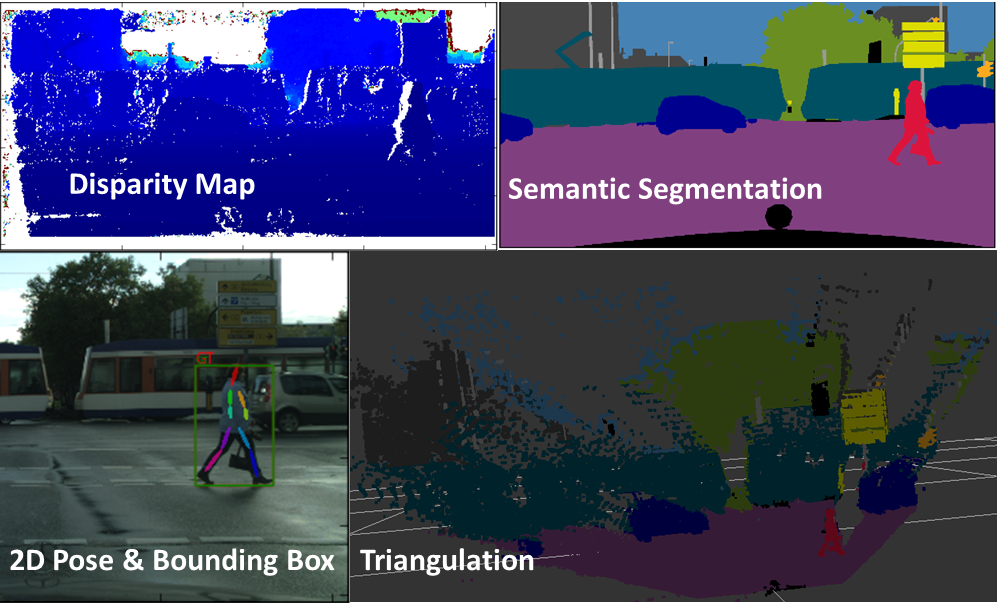}
% \label{fig:triang}
% \subfigure[Segmentation]{
% \includegraphics[width=.4\textwidth]{figures/Triangulation/image32.png}
% }
% \subfigure[Disparity map]{
% \includegraphics[width=.4\textwidth]{figures/Triangulation/image19.png}
% }
% \subfigure[3D reconstruction]{
% \includegraphics[width=.6\textwidth]{figures/Triangulation/image36.png}
% }
% \subfigure[3D reconstruction]{
% \includegraphics[width=.3\textwidth]{figures/Triangulation/darmstadt_8_pose.png}
% }
\vspace{-0.8cm}
\caption{A scene's semantic segmentation, disparity map, triangulation of the frame from the disparity map and the triangulated human pose.}\label{fig:triang}
\end{figure}
\subsection{Reconstructing Pedestrians} 
\begin{figure}[h]
\centering
 \includegraphics[width=\textwidth]{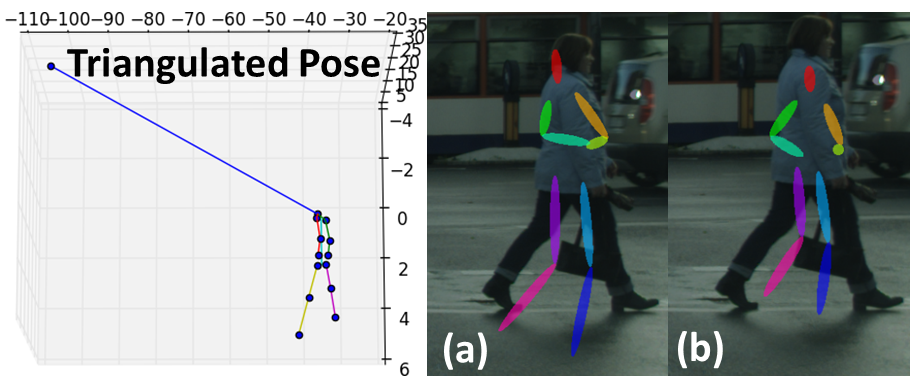}
 %\vspace{-0.8cm}
\caption{\emph{To the Left:} Incorrectly triangulated head position, full image in \Figure{fig:triang}. All axis are in meters. Procrustes corrected skeletons with (a) scaled according to backbone length and (b) scaled according to hipbone length. Neither of the scalings give the desired result.}\label{fig:3DPoseReprojection}

%  <---------------------------------Crop these images
% \subfigure[Skeleton scaled according to backbone length]{
% \includegraphics[width=.4\textwidth]{figures/3DPoseCorrection/image24.png}
% }
% \subfigure[Skeleton scaled according to hipbone length]{
% \includegraphics[width=.4\textwidth]{figures/3DPoseCorrection/image40.png}
% }
 \vspace{-0.5cm}
\end{figure}

Triangulation can be used to reconstruct the human 3D poses from the 2D poses found with the dataset's disparity maps. This however results in a noisy pose estimate as seen in \Figure{fig:triang}. Stereo triangulation results in a noisy 3D reconstruction. The triangulated 2D joint positions can therefore receive incorrect depth resulting in implausible 3D poses. Often a body joint receives the depth of the background resulting in an elongated limb, as seen in \Figure{fig:3DPoseReprojection}. 
To correct such errors we apply a threshold to limb lengths, proportioned according to the hip length or back-bone length of the pedestrian. This is not robust because the hip and backbone length are estimated according to a standard skeleton from Human3.6M\cite{ionescu2013human3}. The limb length can be estimated according to an average skeleton relative to the height of a person. The height of the person can be roughly approximated from the bounding box height, with the downside that BBox height is pose-dependent. 

The corrected skeleton may still suggest a physically implausible pose. To correct this the nearest neighbour plausible pose is found from Human3.6M\cite{ionescu2013human3}. 
To find an outlier robust estimate of the nearest neighbor a thresholded loss is applied. Procrusets analysis is used to find the optimal alignment between the skeletons. The final corrected pose with scaling according to hip or backbone are shown in \Figure{fig:3DPoseReprojection}.
\begin{figure}[h]
\centering
 \includegraphics[width=\textwidth]{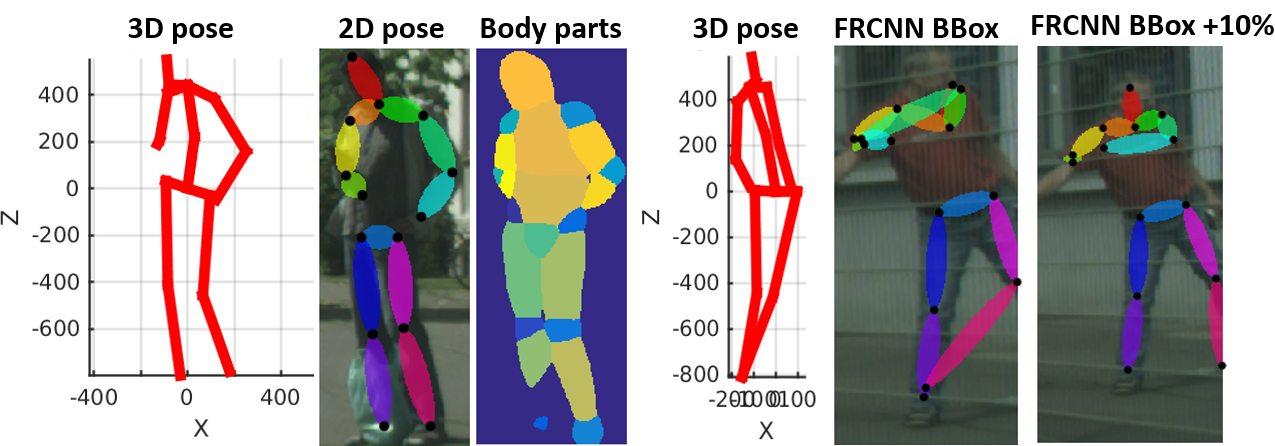}
 \vspace{-0.8cm}
\caption{To the \emph{Left} \emph{DMHS} estimates the pose of a pedestrian decently correctly when the pedestrian is clearly visible. The 3D pose (scale in cm), 2D pose and Body part segmentation are shown respectively. To the \emph{right} BBox enlargement improves the pose estimation when some limbs are not visible.}\label{fig:dmhs}
\end{figure}

It is clear that the scaling and rotation of the resulting 3D pose are imperfect. When triangulating the pose for each frame jitter can be expected between frames due to noise. Therefore a monocular single-person 3D pose estimator \emph{Deep Multitask Architecture for Fully Automatic 2D and 3D Human Sensing }(DMHS)\cite{popa2017deep} is tested as well.

The \emph{DMHS} is applied to GT and \emph{FRCNN} see \Figure{fig:dmhs} \emph{left} and \emph{right} respectively. At times \emph{FRCNN} provides a too small BBox, by increasing the boundary (with 10\%) the results improve, see \Figure{fig:dmhs} \emph{right}.
The pose detector fails when
multiple people are present in the BB, or when the pedestrians are poorly visible.  Eventhough \emph{FRCNN} detects close-up pedestrians, still very few BBoxes are clearly visible and thus few obtain accurate pose estimations. 
\subsection{Reconstructing Vehicles}
\begin{figure}[h]
\centering

 \includegraphics[width=\textwidth]{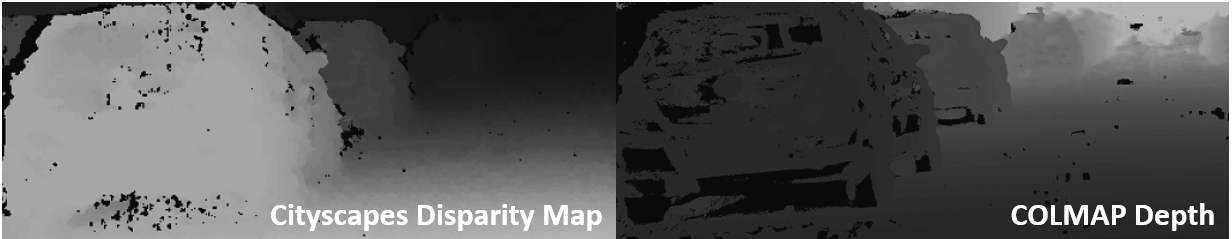}
 \vspace{-0.5cm}
 \caption{To the \emph{left} Cityscapes disparity map and \emph{right} COLMAP depth estimate of a \emph{FRCNN} BBOx of a car clearly contain multiple cars.}\label{fig:3Dvehicle}
% \caption{Depth maps of vehciles in a FRCNN BB.}
% %  <---------------------------------Crop these images
% \subfigure[Colmap's estimated depth for one car's BBox ]{
% \includegraphics[width=.4\textwidth]{figures/Reconstructing_vehicles/car_1.png}
% }
% \subfigure[Cityscape's estimated depth for one car's BBox ]{
% \includegraphics[width=.4\textwidth]{figures/Reconstructing_vehicles/car_2.png}
% }
% \subfigure[Top view image of the final reconstrcution containing estimated cars in black from multiple frames. The car road is in purple and the sidewalk in pink.]{
% \includegraphics[width=.6\textwidth]{figures/Reconstructing_vehicles/scene.png}
% }
\end{figure}
\begin{figure}[h]
\centering
\label{fig:pannetvehicle}
\caption{PANNET correctly separates different parked cars even in the presence of occlusions.}
%  <---------------------------------Crop these images
 \vspace{-0.3cm}
\includegraphics[width=.6\textwidth]{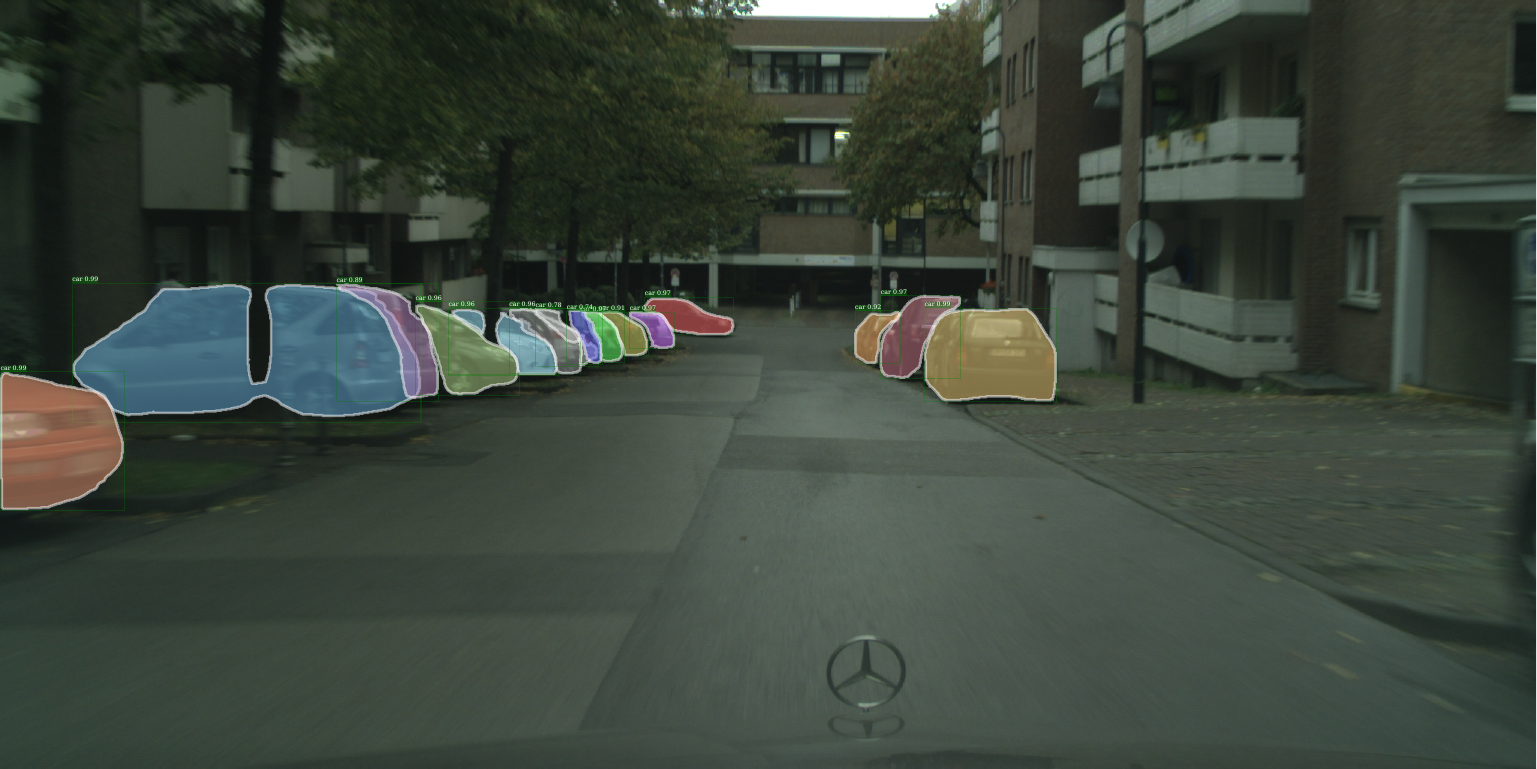}
\end{figure}

 The \emph{FRCNN} has trouble separating multiple instances of cars when cars are parked in a row as seen in \Figure{fig:3Dvehicle} due to the large visual overlap. During triangulation for simplicity we model cars found by the detection model by fixed sized 3D BBoxes. As a result during 3D triangulation multiple vehicles that are visible in one 2D BBox get replaced by one 3D BBox with an average disparity for all ofthe cars visible in the 2D bbox. This results in an incorrect 3D reconstruction of the scene. To improve this Path Aggregation Network for Instance Segmentation (PANNET)\cite{liu2018path} an \emph{FRCNN} architecture based instance segmentation network is utilized instead. Sample segmentation showing the correct instance segmentation of \emph{PANNET} is shown in \Figure{fig:pannetvehicle}.

\section{Developments in the field}
The presented results were developed from 2016 to 2018. In parallel to our work \cite{kong2022research,feng2023deep} noted that detecting objects at a distance is hard, and in particular human detection at a distance or in the presence of occlusions has gained popularity\cite{huan2023mlffcsp,cheng2019efficiently,liu2019improving,cao2019taking,xu2020beta,liu2020coupled,hagn2022validation,zou2023correlation}. Further it has been noted by \cite{Hasan_2021_CVPR} that a number of human detection models do not generalize across datasets. Optimal alignment of BBoxes has also been studied in \cite{madan2023drawing}. The run-time accuracy trade-off of object detection methods aimed to be utilized on AVs is studied in\cite{wang2023we}.

More compact representations of scenes are often utilized in AV planning containing either rasterized graphs with local context\cite{chou2020predicting} or BEV representations\cite{li2022bevsurvey}.  This is suitable as planning must occur fast, but we still believe that articulated human motion ought to be included in the representation. The advantage of utilizing 3D pointclouds and images is that the 3D reconstructed scenes can easily be utilized to train AVs on augmented data\cite{Priisalu_2020_ACCV, Priisalu_2021_CoRL, priisalu2023varied}. 

Human and object detection in traffic from alternative sensors such as Radar\cite{yahia2023radar}, LiDAR\cite{chen2023voxelnext,zhang2020stinet,li2020deep,mseg3d_cvpr2023,sdseg3d_eccv2022,Sun2021RSNRS}, event cameras\cite{colonnier2022event} have been studied as a way to boost object and human detection performance. Methods to improve low-quality image data by reducing motion blur\cite{torres2023depth}, increasing image quality in low light or low-resolution images \cite{aakeberg2023relief} performing detection on RAW images\cite{ljungbergh2023raw}, or object in-painting to recover from occlusions\cite{lugmayr2022repaint} could possibly greatly improve human sensing in real traffic data. 

Further LiDAR sensors have gained popularity to avoid the difficult task of estimating the depths of moving objects. Unfortunately sensing of articulated humans in LiDAR\cite{yang2021s3} has not yet caught up with the methods developed to sense humans in videos. There exist methods that combine LiDAR and RGB fusion\cite{fadadu2022multi,li2020end,cui2021deep,alfred2023fully,Zanfir2022HUM3DILSM} for pedestrian detection and trajectory forecasting, the same could be done for human pose forecasting. Human pose estimation has developed greatly with more models that fit meshes to human bodies to densely estimate human pose and semantic mask\cite{xu2020ghum,popa2017deep,Saini2022SmartMocapJE,yang2021s3,SPEC_ICCV_2021}, methods that reason about the physics of the estimated pose have been developed\cite{Gartner_2022_CVPR,Gartner_2022_CVPR2,Tripathi20233DHP}, as well as methods that utilize temporal constraints\cite{xu20hieve,chang20towards,yuan2020simple}. Still, the majority of articulated human sensing methods are developed on visuals where humans are centered in the images\cite{xu2020ghum,popa2017deep,Gartner_2022_CVPR,Gartner_2022_CVPR2,Tripathi20233DHP,Saini2022SmartMocapJE}, leaving a gap to traffic data where most humans are relatively far away. Human pose estimation and forecasting are more frequently being combined with segmentation\cite{deng2923split,Das2022DeepMN,Kishore2019ClueNetA}, tracking\cite{deng2923split,RAJS2023104632,you2021multitarget} gait recognition\cite{MengHTL23} and camera pose estimation\cite{Saini23smart,ye2023slahmr}. Human motion can be very informative in traffic and pedestrian behavior modeling can even be used to detect vehicles in blind spots in traffic \cite{Hara2020predicting}.
To maximally utilize the available information in human motion, methods that are robust to variations in the human physique and behavior need to be developed, but this is hard due to the relative lack of data. 

Vehicle orientation and shape estimation techniques are amongst others in images\cite{shi2021optimal}, in LiDAR\cite{goforth2020joint}. A method that jointly performs semantic segmentation and 3D reconstructions could benefit both tasks\cite{hayakawa2020recognition}.  

Improved depth estimation of pedestrians and vehicles through a LiDAR data-like data representation Pseudo-LiDAR is studied in \cite{you2019pseudo,wang2019pseudo}
Finally, 3D reconstruction methods have developed greatly with more learning integrated into the 3D reconstruction pipeline \cite{wei2020deepsfm,xiao2023level,Zhu_2022_CVPR,Sarlin_2021_CVPR,zhu2023nicer}, from learned monocular depthmaps\cite{zhao2020monocular}, to learned 3D reconstruction features \cite{detone2018superpoint} to 3d matching\cite{sarlin2020superglue} and the visually pleasing NERF-based methods\cite{mildenhall2020nerf,li2022read}. Semantic segmentation, tracking, and object detection methods are also becoming less supervised utilizing learned matching and language model based labels \cite{li2022elevater,ovtrack,rts,zhou2022survey,zou2023segment,zhang2023simple,li2023semantic,zou2023generalized,harley2020tracking}. Combining different visual tasks like object detection semantics segmentation, tracking with 3D modeling has seen success in \cite{harley2020tracking,vaquero2017deconvolutional}. This is quite natural because as seen in \Figure{fig:3Dvehicle} the two tasks are closely intertwined and information sharing may help in both directions. 

Traffic datasets that are focused on pedestrians have become more abundant \cite{ceaser2020nuscenes,PhamSPZPCM0020,peishan2022stcrowd,ZhangXWXLG20,BreitensteinF22,WangC0LSLL20,NeumannKZSPMPTV18,RasouliKKT19,HudaHGM20,Devansh2022comparison,XuZLZP18,tumas20pedestrian}, but there exist only datasets with estimated articulated labels for pedestrians \cite{windbacher3dpose}. Even though progress has been made on marker-less human motion capture \cite{von2018recovering} the methods need to be made robust for multiple humans at a distance and in the presence of occlusions. In parallel to our work, a study\cite{ZhangBS17} on occlusion rates in pedestrian bounding boxes on the Cityscapes dataset was performed. We note that \cite{ZhangBS17} may be treated as complementary to the work presented here that focuses on the task of 3D human pose reconstruction rather than just bounding box occlusions.

\section{Conclusion}
None of the discussed methods of 3D reconstructing human pose are robust enough to be utilized to forecast human motion for assisted driving. This is because there is a gap in performance for human sensing methods between the datasets used in standard benchmarks and the performance on real traffic data, suggesting that benchmarks of human motion sensing are not representative of utilization in traffic. Instead, traffic-based articulated 3D human sensing benchmarks should be developed. 
Available 3D human pose datasets in the wild \cite{von2018recovering} still lack in distant pedestrians under poor lighting conditions, or provide only approximate human poses\cite{windbacher3dpose}. 
To make articulated human sensing robust temporal smoothness, consistent use of an individual's estimated limb lengths, foreseeing typical motions given the human's environment, and understanding of the physical constraints of the human body should be solved simultaneously as the problems share information. So far a number of methods have solved some of these subproblems, but a unifying method is still to be developed. As a result to the lack of a robust articulated humans sensing method a large number of existing autonomous vehicle planning models\cite{bae2023set,shen2018transferable,li2020learning,rhinehart2019precog,yao2021bitrap,huang2020long,deo2020trajectory,zhao2021tnt,mangalam2020not,li2022evolvehypergraph,chen2022hierarchical,djuric2020uncertainty,liang2020pnpnet,luo2022gamma,ma2019trafficpredict,zhu2020robust,sriram2020smart,zhao2019multi,luo2021safety,fang2020tpnet,park2020diverse,vansafecritic,yang2020traffic,cheng2021amenet,giuliari2021transformer,AndersonVJ20a,SalzmannICP20,HamandiDF19,Yao2019FollowingSG,ChenLSL20,IvanovicLSP21,GirgisGCWDKHP22,li2020end,chou2020predicting} treat pedestrians by their bounding boxes, thus omitting the motion cues
available in human pose and therefore ignoring available future motion cues. If robust and complete articulated human sensing methods are developed, then complete human forecasting methods may be developed and utilized in the planning stages of AVs.
\bibliographystyle{unsrt}
\bibliography{main}

\appendix
\section{3D reconstruction systems overview}
Open Structure for Motion Library (OpenSFM)\cite{Mapillary}- A Structure for Motion system, that is an incremental 3d reconstruction system. Uses Hessian Affine Feature Point detector\cite{Mikolajczyk04scale} and Histogram of Oriented Gradients (HOG)\cite{mcconell86hog} descriptors jointly which are nearest neighbor matched\cite{Muja2009FastAN} across images. A rotation-only transformation is found between the first frames if at least 30$\%$ of the points are outliers (to ensure a large enough change in viewpoint). From the initial pair, a sparse 3D cloud is found by the 5-point algorithm or by assuming planar motion of the camera, whichever performs best. The resulting camera matrices are then used for triangulation and bundle adjustment (BA). Additional frames are added according to the largest overlap with the existing pointcloud, and are aligned with the pointcloud to minimize the re-projection error of the pointcloud. BA is applied after adding new images to the pointcloud.
    Fails to reconstruct the Cityscapes scenes likely because the change in camera rotation is too small between frames.
    
    Bundler\cite{snavely2006photo} is also a SFM system. It detects SIFT\cite{lowe2004sift} features that are matched with approximate nearest neighbors\cite{arya1998optimal}. RANSAC\cite{fischler1981random} is used to find the fundamental matrix with the 8-point algorithm\cite{hartley2003multiple}. The fundamental matrix is refined, its outliers are removed and keypoints tracks are checked for consistency.  Levenberg-Marquardt\cite{nocedal1999numerical} is used to find the first camera matrices and in sparse bundle adjustment\cite{lourakis2004design} for any additional cameras (images chosen in the order of largest number of matches with triangulated points) that are initialized with Direct Linear Transform (DLT)\cite{hartley2003multiple}. The initial image pair is chosen such that there is a large enough rotational difference between the images. Finds <10 matches, and fails again likely because the images are blurry and the rotational difference between the camera views is too small.
    
    OpenCV Structure from motion library\cite{opencv} - A SFM library that uses DAISY features\cite{tola2009daisy}. Finds the essential matrix with RANSAC and the 8-point algorithm. And an inexact Newton method Schur-based solvers\cite{agarwal2010bundle} to optimize BA. Result of 30 frames - finds relatively few points without clear structure. See \Figure{fig:opencvBA}
     
     VisualSFM\cite{wu2013towards} a parallelized SFM pipeline with Bundler. Uses SIFT on the GPU\cite{wu2011siftgpu} and Multicore Bundle adjustment\cite{wu2011multicore}. Only a thresholded number of large-scale features are matched across images.
     This unfortunately fails possibly because of image blur or the lack of distinct large-scale structures in the images. The method is unable to find enough SIFT feature points likely because the images are blurry and finds no verified matches between two stereo images. Finally, VisualSFM cannot handle forward motion, not finding a good initial pair of images with enough matches. 
     
 ORBSLAM\cite{murartla2017orb}- ORB-feature\cite{rublee2011orb} (a fast feature descriptor combining gradient and binary features) based Simultaneous Localization And Mapping system. ORB features from the left image are matched along epipolar lines with ORB features from the right image, and the disparity is calculated. Points that are further than 40 baselines away from the camera are ignored. The points that are close to the camera are triangulated, and the left camera is considered to be the origin. Additional cameras are added by performing camera position optimization in BA between matched 3D points and keypoints in the new frame, BA of the newly added added keypoints, and by finally performing full BA after loop-closure detection and correction. BA is optimized with Levenberg–
Marquardt implemented as g20\cite{kummerle2011g} Finds too few keypoints, likely due to blur and depth threshold. Results in a too sparse reconstruction.

COLMAP\cite{schoenberger2016sfm,schoenberger2016mvs}- an incremental Structure for Motion and Muti-view stereo system. Extracts SIFT\cite{lowe2004sift} features that are exhaustively matched (other matching methods are also available) across all images. The reconstruction is built from an initial pair of images, chosen by \cite{noah2008scene,beder2006determining} additional views are chosen by high inlier ratios that approximate uncertainty estimates and by prioritizing images with uniformly distributed keypoints that match with triangulated points. The method uses a robust RANSAC-based triangulation with adaptive outlier thresholds to add new camera views to the reconstruction. Local BA is performed after a new camera view is added. Finally, a global BA is performed iteratively followed by filtering of outliers and degenerate camera view, and triangulation until the BA converges. BA is performed by Preconditioned Conjugate
Gradient\cite{agarwal2010bundle,wu2011multicore} for a large number of cameras and by \cite{davis2005cholmod,lourakis2009sba} for smaller systems. Converges for 150 scenes on the training and validation set and 150 scenes on the test set out of the 3475 scenes available. See \Figure{fig:sparse}.
\section{Additional Results}
\subsection{Error distribution per human body joint}
\begin{figure}[h]
%\vspace{-3cm}
\centering
\label{fig:per_joint_error}
 \includegraphics[width=\textwidth]{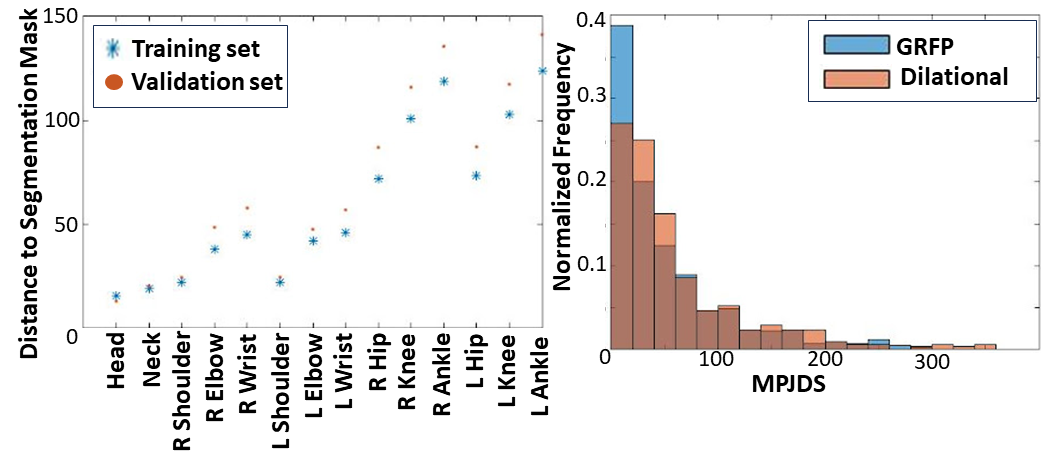}
\vspace{-0.8cm}
\caption{\emph{Left:} The JDS for different joints. The feet are the most difficult to detect. \emph{Right:} Comparison of the MPJDS rates of the Dilational and GRFP net. GRFP has lower MPJDS than the Dilational net.}
\end{figure}

In \Figure{fig:per_joint_error} it can be seen that in general feet are the hardest to estimate the position of, while the head is the easiest. If the articulated human motion is to be predicted we however need the feet position to be accurate to foresee the pedestrian's future velocity.  To improve this temporally smooth human detection and pose estimation methods should be utilized in the future. In \Figure{fig:per_joint_error}\emph{Right} a comparison between the error distribution of \emph{Dilational} net and \emph{GRFP} is shown. It is clear that \emph{GRFP} produces joint estimations that have a lower distance to the human mask.
\subsection{Procrustes Analysis}
\begin{figure}[h]
\centering
\caption{Skeleton with thresholded limb lengths after scaling according to hip bone length shown with its nearest neighbor from the Human3.6M dataset. The longer skeleton is the thresholded 3D reconstructed human pose. The unthresholded skeleton is shown in \Figure{fig:3DPoseReprojection} \emph{Left}.}
\includegraphics[width=0.8\textwidth]{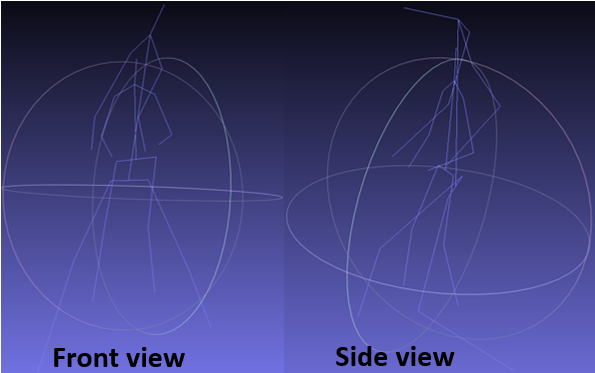}\label{fig:3DPoseCorrection}
% \subfigure[3D Pose of figure on prev. slide after triangulation]{
% \includegraphics[width=.3\textwidth]{figures/3DPoseCorrection/cropped_pose.png}
% }
% \subfigure[3D Pose of triangulation and the closest neighbour after Procrustes analysis front view]{
% \includegraphics[width=.3\textwidth]{figures/3DPoseCorrection/cropped37.png}
% }
% % <--------------------------- Add skeleton of nearest neighbour without Procrustes.
% \subfigure[3D Pose of triangulation and the closest neighbour after Procrustes analysis side view ]{
% \includegraphics[width=.3\textwidth]{figures/3DPoseCorrection/cropped_image.png}\label{fig:3DPoseCorrection}
% }
\end{figure}

The results of aligning the thresholded pose with its nearest neighbor from Human3.6M is shown in \Figure{fig:3DPoseCorrection}. Because all limb lengths are reconstructed under noise the different limbs get elongated or compressed to a different degree. Therefore scaling the pose according to hip length results in this case in a too large scaling factor because the hip length is compressed in the reconstruction. The backbone is elongated so scaling according to the backbone results in a too small scaling factor. The elongations and compression of the different limbs vary from one triangulated pose to another, making the choice of a scaling factor hard. The height difference between the feet and the heat depends on the pose and is therefore not a suitable measure for scale. The same applies to the distance between the feet and hands.
\subsection{Additional Pedestrian Detection and Pose Estimation}
\begin{figure}[h]
\centering
\includegraphics[width=\textwidth]{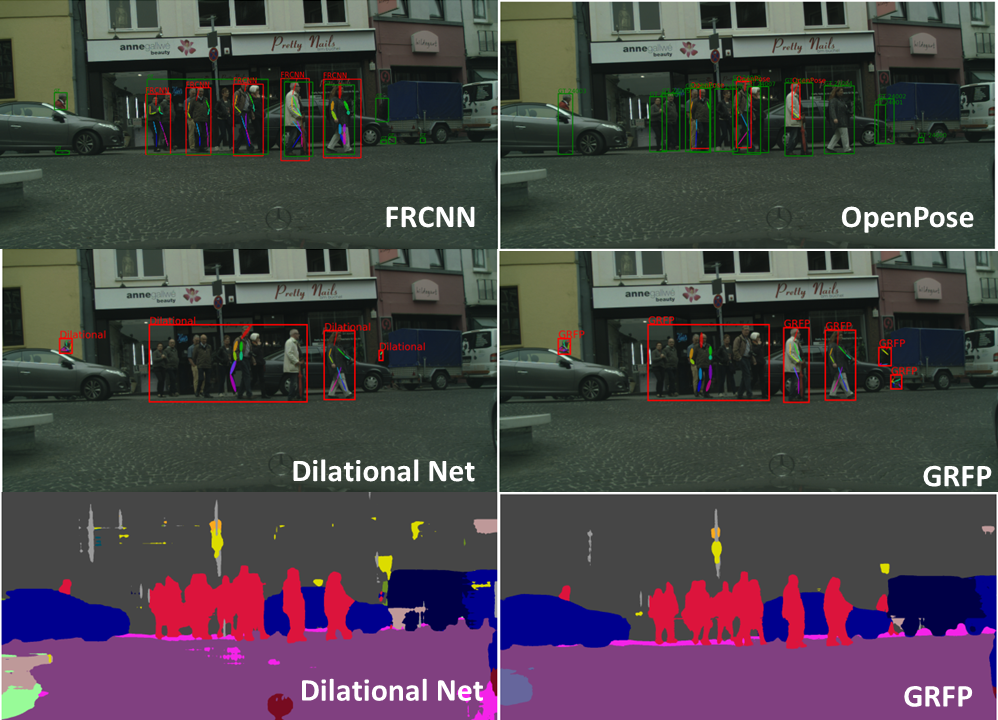}
\caption{OpenPose when applied on the whole image detects only a few pedestrians. FRCNN detects some selected pedestrians. The segmentation networks can detect all of the pedestrians, but because they produce only class labels one single BBox is given to multiple pedestrians. GRFP has smoother segmentation than DilationalNet and results in a better separation of the pedestrian Bboxes.}\label{fig:seg3}
\end{figure}
\begin{figure}[h]
\centering

\includegraphics[width=\textwidth]{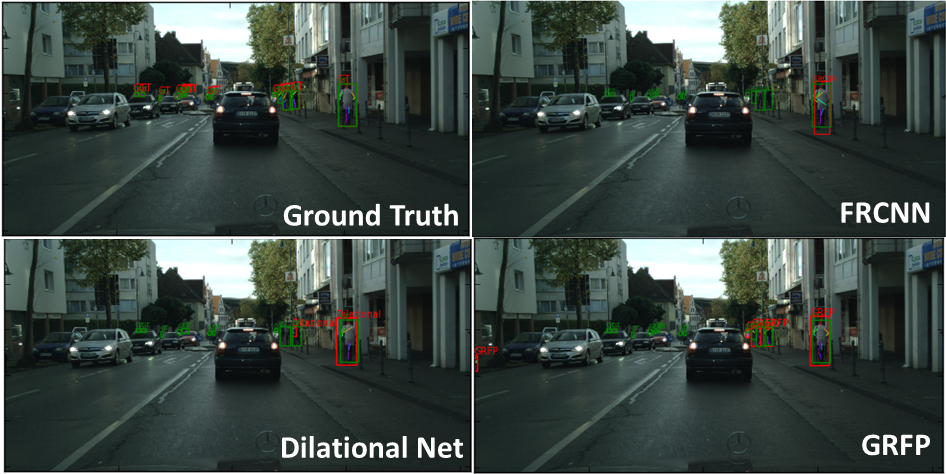}
\caption{FRCNN misses a large number of the distant pedestrians. The Dilational net and GRFP detect more distant pedestrians than FRCNN and GRFP results in a more accurate 2D pose for the closest pedestrian to the right.}\label{fig:seg2}
\end{figure}
\begin{figure}[h]
\centering
\includegraphics[width=\textwidth]{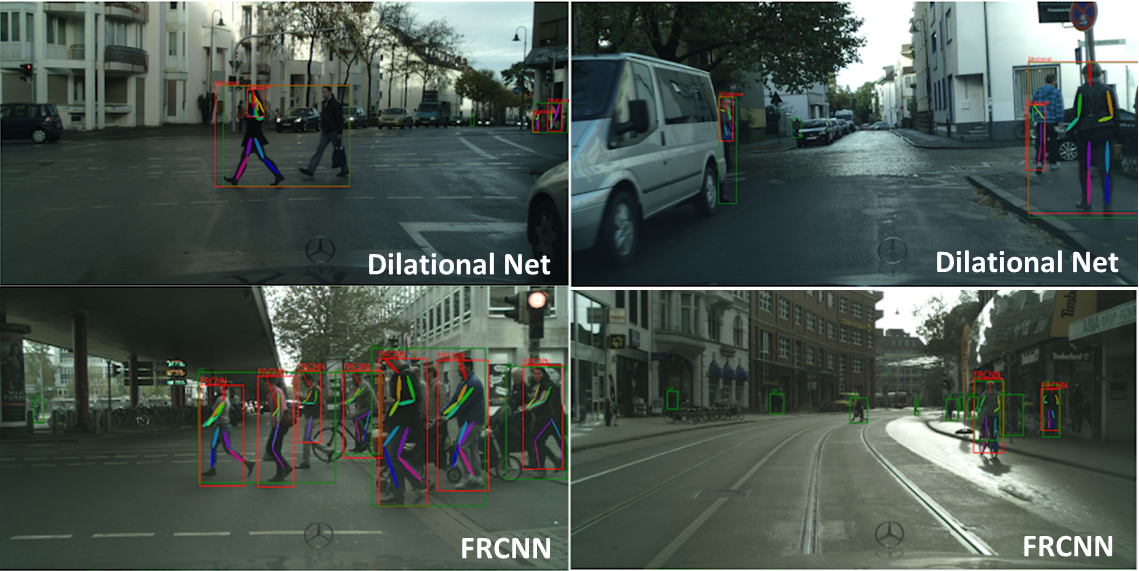}
\caption{On the top: Dilational Net can detect pedestrians even in the presence of occlusions, but produces one single bounding box for close by pedestrians. FRCNN can detect pedestrians that are close to the camera well, but fails to detect far away pedestrians.}\label{fig:seg4}
\end{figure}
\begin{figure}[h]
\centering

\includegraphics[width=\textwidth]{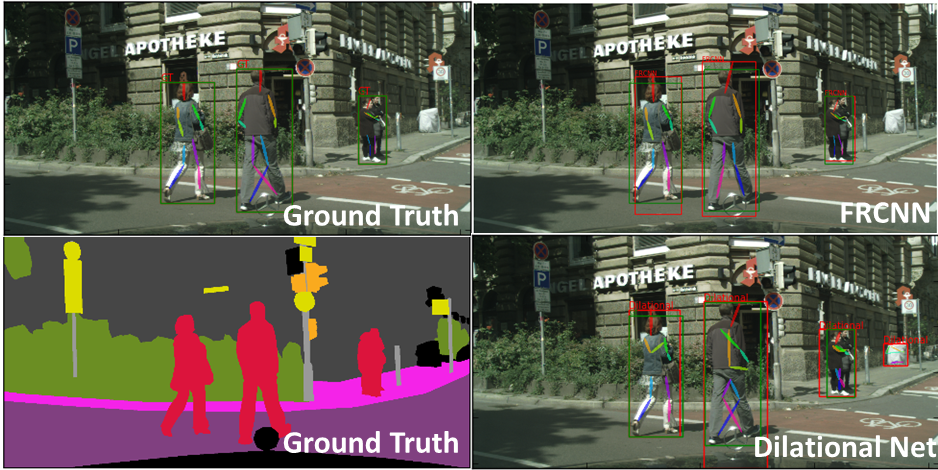}
\caption{The placement of the bounding box affects the estimated 2D body pose. FRCNN produces larger bounding boxes than found by the GT segmentation mask. This produces a more correct 2D body pose.}\label{fig:seg1}
\end{figure}

OpenPose misses a large number of pedestrians (with variations from frame to frame) as seen in the supplementary video at \url{https://youtu.be/qpxpdtHbbGA} and \Figure{fig:seg3}. OpenPose misses pedestrians due to large distance to camera, poor lightning and motion blur as seen in the supplementary video. OpenPose can produce impressive results when pedestrians are close to the camera but at a distance it has trouble with occlusions and can produce a number of odd false positive pedestrian detections, as seen in the video. Pedestrian detection is improved by applying FRCNN object detector. But FRCNN still omits distant pedestrians as seen in the video and \Figure{fig:seg3},\Figure{fig:seg2}, and \Figure{fig:seg4}. On the other hand the semantic sgementation networks are  susceptible for false positive as seen in the video. Finally it can be noted that on close by pedestrians FRCNN produces bboxes that are larger than the pedestrian often leading to better 2D pose estimates than the segmentation network, as seen in \Figure{fig:seg1}.

\subsection{Additional DMHS results}
\begin{figure}[h]
%\vspace{-3cm}
\centering
 \includegraphics[width=\textwidth]{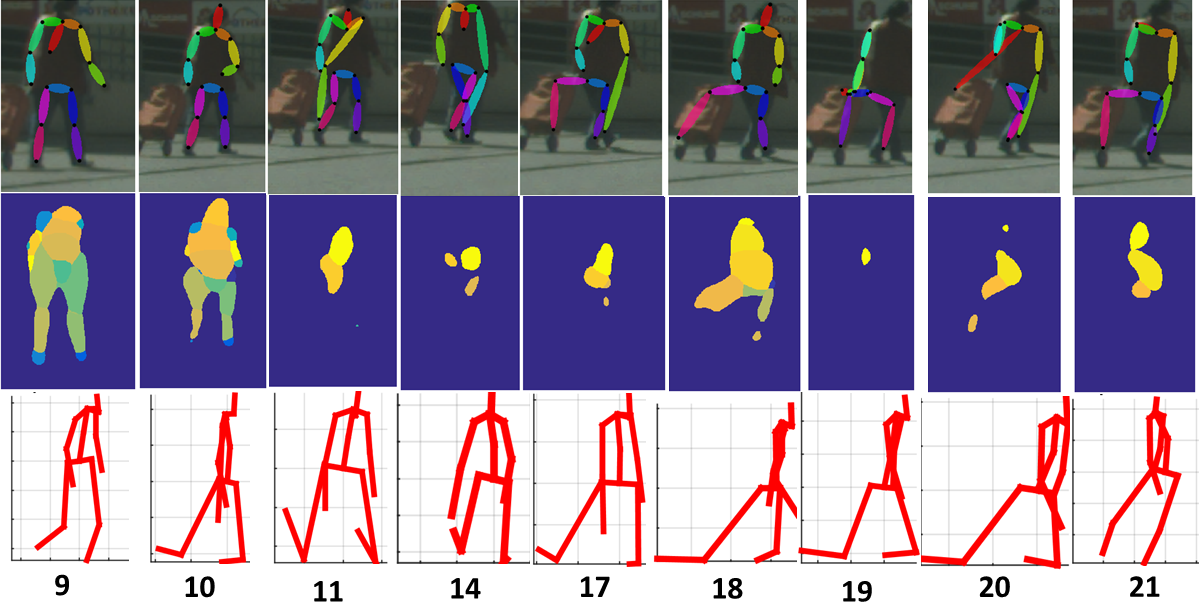}
\vspace{-0.8cm}
\caption{Consequtive pedestrian detections by FRCNN followed by 2D pose(top row), body parts segmentation (middle row) and 3D pose estimates(bottom row). No pedestrian is detected in frames 12,13 and 15,16. The FRCNN BBoxes jump around the pose estimates to jump. When the pedestrians head is not visible then the body part segmentation fails. The 3D poses do not resemble the true 3D pose as the human appears to be crawling on knees in 3D poses. The 2D poses jump from frame to frame.}\label{fig:dmhi_results2}
\end{figure}

\begin{figure}[h]
%\vspace{-3cm}
\centering
 \includegraphics[width=0.9\textwidth]{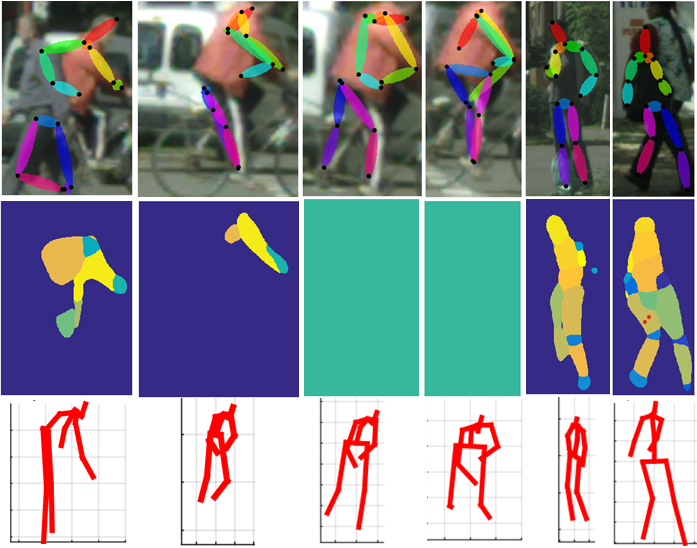}
\vspace{-0.5cm}
\caption{Some varied results of DMHS.DMHS gets confused in the case of multiple occluding humans. And appears to have trouble with body part segmentation when a human is on a bike. The 2D body pose estimate seems to be greatly affected by the poorly fitting FRCNN bboxes that leave out the pedestrians head.}\label{fig:dmhi_varied}
\end{figure}

The DMHS's accuracy is like OpenPose, depent on the bbox placement. Because the FRCNN produces bboxes that jump from frame to frame as seen in \Figure{fig:dmhi_results2}. FRCNN can even jump frames, by being unable to detect pedestrians at some frames. This motivates our suggestion that temporally smoothed methods should be developed for articulated pedestrian detection. DMHS's quality varies from image to image, some samples with quality variations are shown in \Figure{fig:dmhi_varied} 
\end{document}